\documentclass[runningheads]{llncs}

\usepackage{eccv}

\usepackage{eccvabbrv}

\usepackage{graphicx}
\usepackage{booktabs}
\usepackage{wrapfig}

\usepackage[accsupp]{axessibility}  %

\usepackage[pagebackref,breaklinks,colorlinks]{hyperref}
\usepackage{hyperref}

\usepackage{orcidlink}

\usepackage{graphicx} %
\usepackage{amsmath}
\usepackage{amssymb}
\usepackage{amsfonts}
\usepackage{amstext}
\usepackage{multirow}
\usepackage{scalerel}

\usepackage{booktabs}
\usepackage{makecell}
\usepackage{graphicx}
\usepackage{multirow}
\usepackage{float}
\usepackage{caption}
\usepackage{multicol}
\usepackage{placeins}
\usepackage{tabu}
\usepackage{color, colortbl}
\definecolor{Gray}{gray}{0.9}
\definecolor{dGray}{gray}{0.6}
\definecolor{OliveGreen}{rgb}{0,0.6,0}
\definecolor{deepPurple}{rgb}{0.4,0.2,0.8}
\definecolor{lightGreen}{rgb}{0.8, 0.88, .76}
\definecolor{lightBlue}{rgb}{0.82, 0.86, .93}
\usepackage{amssymb} %
\usepackage{pifont}  %
\newcommand{\cmark}{\ding{51}}%
\usepackage{xcolor}

\usepackage[normalem]{ulem}
\useunder{\uline}{\ul}{}

\newcommand{\model}{VisFocus }

\newcommand{\AlgoName}{VisFocus }
\newcommand{\AlgoNameNoSpace}{VisFocus}
\newcommand{\Vilma}{ViLMA }
\newcommand{\VilmaNoSpace}{ViLMA}
\newcommand{\comment}[1]{}

\usepackage[symbol]{footmisc}

\newcommand{\gcol}[1]{{\bf \fontsize{6.5}{42}\selectfont \color{OliveGreen}~(#1)}}
\newcommand{\rcol}[1]{{\bf \fontsize{6.5}{42}\selectfont \color{red}~(#1)}}

\newcommand\blfootnote[1]{%
  \begingroup
  \renewcommand\thefootnote{}\footnote{#1}%
  \addtocounter{footnote}{-1}%
  \endgroup
}

\begin{document}

\title{VisFocus: Prompt-Guided Vision Encoders for OCR-Free Dense Document Understanding}

\author{
    Ofir Abramovich \inst{1}\textsuperscript{*} \and
    Niv Nayman \inst{2}\textsuperscript{\textdagger} \and
    Sharon Fogel\textsuperscript{*} \and
    Inbal Lavi\textsuperscript{*} \and
    Ron Litman\inst{2} \and
    Shahar Tsiper\inst{2} \and
    Royee Tichauer\inst{2} \and
    Srikar Appalaraju\inst{2} \and
    Shai Mazor\inst{2} \and
    R. Manmatha\inst{2}
}
\institute{\textsuperscript{1} Reichman University, Israel
        \,\,\,\,\,\,\,\,
        \textsuperscript{2} AWS AI Labs
}

\titlerunning{VisFocus}

\authorrunning{O.~Abramovich et al.}

\maketitle

\begin{abstract}
In recent years, notable advancements have been made in the domain of visual document understanding, with the prevailing architecture comprising a cascade of vision and language models. The text component can either be extracted explicitly with the use of external OCR models in OCR-based approaches, or alternatively, the vision model can be endowed with reading capabilities in OCR-free approaches.  Typically, the queries to the model are input exclusively to the language component, necessitating the visual features to encompass the entire document. In this paper, we present \textit{VisFocus}, an OCR-free method designed to better exploit the vision encoder's capacity by coupling it directly with the language prompt. 
To do so, we replace the down-sampling layers with layers that receive the input prompt and allow highlighting relevant parts of the document, while disregarding others. 
We pair the architecture enhancements with a novel pre-training task, using language masking on a snippet of the document text fed to the visual encoder in place of the prompt, to empower the model with focusing capabilities. Consequently, VisFocus learns to allocate its attention to text patches pertinent to the provided prompt.
Our experiments demonstrate that this prompt-guided visual encoding approach significantly improves performance, achieving state-of-the-art results on various benchmarks.
\blfootnote{*Work done during an internship$\backslash$employment at Amazon}%
\blfootnote{\textsuperscript{\textdagger} Corresponding author: \href{mailto:nivnay@amazon.com}{nivnay@amazon.com}

}

\keywords{Document Understanding \and OCR-free Models}
\end{abstract}

\begin{figure}[tb]
\vspace{-4mm}
  \centering
  \caption{\textbf{\AlgoNameNoSpace' key contributions.} The left side of the figure illustrates how \AlgoName enables the vision model to better align visual features to the input prompt; Unlike previous approaches, \AlgoName inputs the prompt not only to the language model, but to the vision encoder as well (top left vs top middle). In addition, a novel pre-training task utilizes the enabled interactions with the prompt to focus the model on specific text patches (bottom middle) instead of the entire text (bottom left). The right side of the figure shows the resulting attention map from \AlgoName illustrating how the model focuses on a specific word taken from the query (`Nursing'). }
    \includegraphics[width=1.0\linewidth]{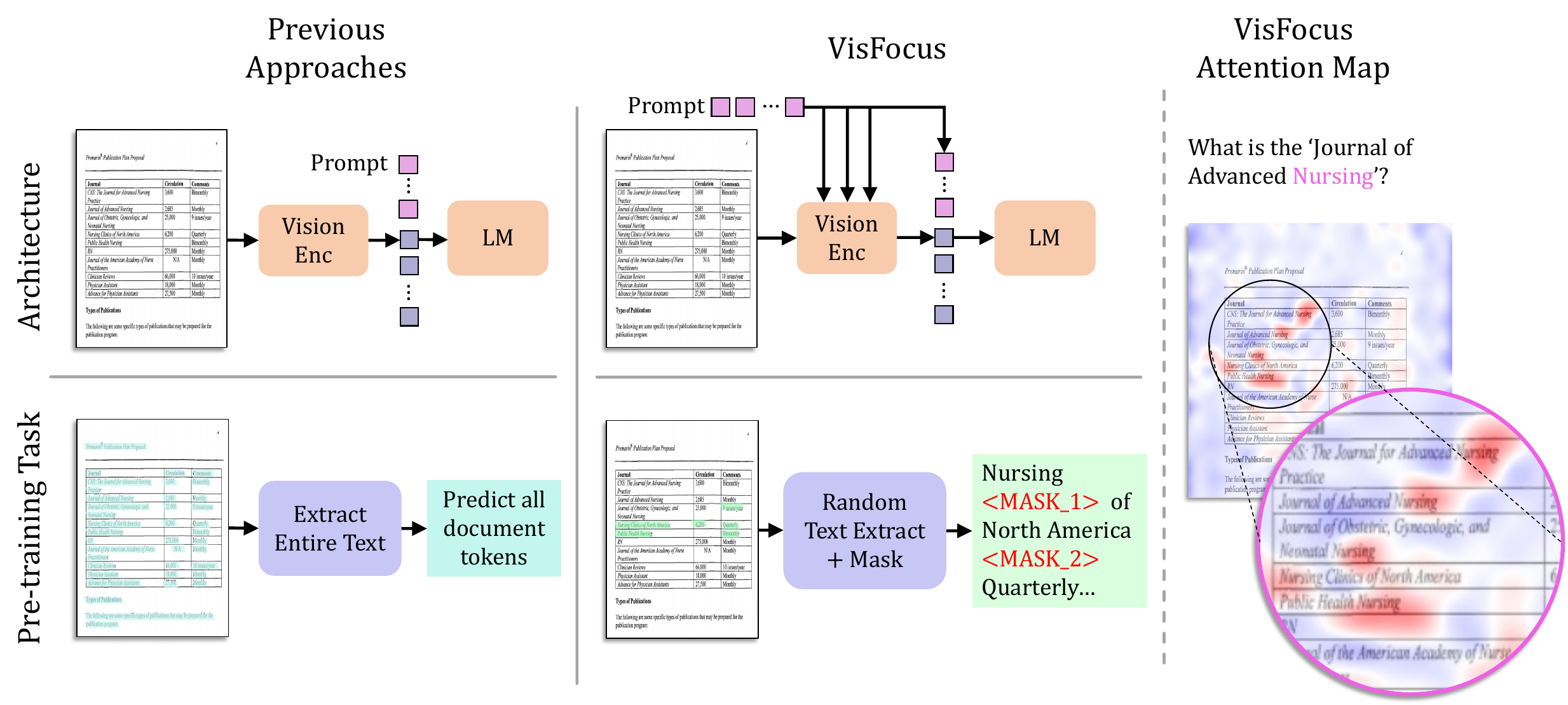}
  \label{fig:teaser}
\vspace{-9mm}
\end{figure}

\section{Introduction}
\label{sec:intro}

Visual Document Understanding (VDU) aims to extract meaningful information from digitized documents, such as PDFs or images, extending beyond the scope of Optical Character Recognition (OCR). This field encompasses various tasks, including DocVQA \cite{mathew2021docvqa}, ChartQA \cite{masry-etal-2022-chartqa}, InfoVQA\cite{mathew2022infographicvqa}, Key-Value Identification in forms \cite{funsdJaume2019}, Entity Extraction \cite{park2019cord} and Document Classification \cite{RVLCDIPHarley2015icdar, M4datasetFujinuma-etal-2023-multi}. 

Traditional VDU approaches rely on OCR to extract textual information from the document\cite{appalaraju2021docformer,xu2020layoutlm,xu-etal-2021-layoutlmv2,huang2022layoutlmv3,appalaraju2023docformerv2AAAI,Tang2022UnifyingVTUDOP,powalski2021going,tanaka2024instructdoc,wang2023docllm}. These Vision-Language (VL)  approaches then take OCR text and spatial features as well as visual tokens of the document as input to generate predictions. However, running OCR at training and inference as a pre-processing step adds additional latency and computational costs \cite{kim2021donut,davis2022dessurt}. In addition, errors originated in the OCR step might propagate to the VL model and deteriorate its performance \cite{taghva2006effects, hwang-etal-2021-cost}.

The OCR-free \cite{kim2021donut, davis2022dessurt, lee2023pix2struct} approach emerged as an alternative way for VDU. Here, the document image is directly fed as input to the vision-language model, bypassing the need for explicit OCR text extraction. The architecture usually consists of a visual encoder, followed by a language model \cite{kim2021donut, davis2022dessurt, lee2023pix2struct, bai2023qwen, ye2023mplug, ye2023docowl, ye2023ureader} which receives as input the visual representation as well as the input query. The model is expected to internally first learn to read, and then perform the down-stream task. To avoid the need for OCR input, an extensive pre-training stage is performed to endow the vision model with reading capabilities \cite{kim2021donut, davis2022dessurt, lee2023pix2struct}. 

In most OCR-free VDU prior art~\cite{kim2021donut, davis2022dessurt}, the user query is used as an input to the language model alone, as illustrated on the top left of \cref{fig:teaser}. Specifically, since the visual features are processed independently of the input language query we posit that the visual features could be sub-optimal, containing information irrelevant to the user query. 
This misalignment is particularly critical for dense documents, that contain a large amount of text. For such documents, reading the text properly requires high-resolution input images, containing many pixels of blank areas, figures and text irrelevant to the user query. 
Intuitively, these can draw a large portion of the encoded visual tokens, while missing sufficient focus on the desired query.

We suggest a new approach for OCR-free VDU, \AlgoNameNoSpace, to generate prompt-aware visual features. This is achieved by (1) incorporating the user prompt directly in the vision-encoder architecture (top middle of \cref{fig:teaser}) and (2) proposing a complimentary pre-training scheme (bottom middle of \cref{fig:teaser}), that through this coupling, enables the prompt to focus the model on the relevant text in the document (right part of \cref{fig:teaser}).
Our approach is inspired by the selective scanning method one might employ when searching for an answer within a document: rather than meticulously reading through every word, attention is focused on identifying keywords pertinent to the question. Once these keywords are identified, closer scrutiny is applied to the surrounding text to extract the desired answer.
Similarly, in \AlgoNameNoSpace, the language prompt assigns more weight to relevant visual features by repeatedly merging visual patches of the input document through designated cross-attention mechanism~\cite{vaswani2017attention} with the input prompt. We term the newly introduced layers \textit{\textbf{Vi}sion-\textbf{L}anguage \textbf{M}erging \textbf{A}ttention} (\textit{ViLMA}) layers. We empirically show that those carefully located \Vilma layers lead to better alignment of visual and language information, enabling the model to focus on contextually relevant information associated with the language prompt, as illustrated in \cref{fig:teaser} (right).

Once the interactions between the textual prompt and visual features are enabled within the model architecture, a complimentary pre-training stage is devised. 
This newly introduced pre-training task, \textit{\textbf{L}ocalized \textbf{M}asked \textbf{P}rompt \textbf{M}odeling} (\textit{LMPM}), leverages these interactions for guiding the model to search for semantically-related text to the prompt rather than reading the entire document. This task is illustrated on the bottom middle of \cref{fig:teaser}. The underlying concept is to enable the model to develop hierarchical understanding of the document, initially learning general reading skills  \cite{kim2021donut, davis2022dessurt} and subsequently refining its ability to focus on specific parts of the text during the second stage.

By carefully combining the \Vilma layers and the the LMPM task, the visual encoder learns to focus its attention on the most relevant patches of the input document. Our empirical analysis shows the contributions of the suggested components (see \cref{ssec:ablation_study}); the LMPM pre-training stage, the ViLMa layers. Thus, exhibiting a symbiosis between those architectural and pre-training enhancements. %
To summarize our key contributions:
\begin{enumerate}
\item We propose novel patch-merging layers, termed \Vilma, 
which imbue the visual encoding process with prompt awareness, resulting in improved alignment between vision and language for VDU tasks. 
\item A new pre-training task tailored for OCR-free VDU, named LMPM, is introduced. 
This task encourages the visual encoder to extract visual features relevant for the specific prompt, enhancing the model's ability to discern relevant information.
\item Through extensive experimentation, we showcase the synergistic impact of combining these architecture enhancements with the introduced pre-training task. This leads to state-of-the-art performance on multiple benchmarks compared to previous similarly sized OCR-Free methods.
\end{enumerate}

\section{Related Work} \label{sec:related}

Document understanding approaches have been widely explored in recent years. In this domain, two primary approaches have emerged: (a) OCR-based approaches, which rely on document OCR to interpret document images, and (b) OCR-free approaches, which bypass OCR to directly solve high-level tasks.

\subsubsection{OCR-based} 
Initially, research in VDU predominantly relied on OCR-based approaches, where a general natural language model is employed alongside spatial features extracted from 2D document images, in conjunction with pre-extracted OCR text\cite{appalaraju2021docformer,appalaraju2023docformerv2AAAI,xu2020layoutlm,xu-etal-2021-layoutlmv2,huang2022layoutlmv3,Tang2022UnifyingVTUDOP,powalski2021going,tanaka2024instructdoc,wang2023docllm,Powalski2021GoingFBTILT}. In recent years, advancements in this field have been notable. Modern methodologies, such as DocFormer~\cite{appalaraju2021docformer,appalaraju2023docformerv2AAAI} and LayoutLM\cite{xu2020layoutlm,xu-etal-2021-layoutlmv2,huang2022layoutlmv3} leverage transformer-based architectures.
These models are specifically designed to integrate spatial features, text tokens, and their corresponding bounding boxes into rich representations, enabling more effective document understanding. 
Additionally, frameworks like UDOP~\cite{Tang2022UnifyingVTUDOP} aim at establishing aligned representations of spatial and textual embeddings, which are then fed into a unified encoder.  This strong reliance on OCR presents several drawbacks: (1) it relies on off-the-shelf tools, making it susceptible to their errors; (2) it increases the complexity of models and computational overhead (3) processing the entire extracted text can result in unnecessary computations, especially in cases where only specific regions or aspects of the document are relevant to the task at hand. This paved the way for OCR-free approaches.

\subsubsection{OCR-free}
Donut~\cite{kim2021donut} and Dessurt~\cite{davis2022dessurt} were pioneering works in the realm of OCR-free  VDU, introducing models that exclusively process document images without reliance on OCR. These works have shown the significance of equipping models with reading capabilities during pre-training to enhance downstream task performance. Subsequently, Pix2Struct~\cite{lee2023pix2struct} demonstrated performance gains by training larger models with expanded datasets and incorporating additional pre-training tasks. 
Notably, Pix2Struct opts for rendering the prompt onto the input image, integrating it visually rather than inputting it directly to the language model. This makes the visual features prompt-dependant; however, the use of rendering limits the semantic usefulness of the input prompt.
In contrast, \AlgoName leverages the prompt in a manner that facilitates semantic understanding, enabling the visual features to prioritize relevant information more effectively.
A concurrent work to ours ~\cite{ganz2024question} injects the user prompt to arbitrary self-attention layers \cite{vaswani2017attention} of a ViT \cite{dosovitskiy2020image} encoder to promote the alignment of visual and lingual features of VL models. While excelling on scene-text images of typically few words, it lacks the crucial complementary pre-training task for reading text segments relevant to the prompt and thus reports low performance for VDU.

A notable category of OCR-free methods is Large Vision-Language Models (LVLMs), including Qwen-VL\cite{bai2023qwenvl}, PaLi-3\cite{chen2023pali3}, MPlug-DocOwl\cite{ye2023docowl}, ScreenAI\cite{baechler2024screenai} and others \cite{liu2023visual, zhang2023llavar, alayrac2022flamingo, li2023blip2, ye2023mplug}. Their main theme is scaling up both the vision encoder, e.g. to ViT-L, ViT-H or ViT-G \cite{DBLP:journals/corr/abs-2010-11929,zhai2021scaling}, and the LM to Large Language Models (LLMs), e.g. ~\cite{qwen, tay2022ul2, touvron2023llama}. Those two components are often connected by advanced alignment modules \cite{liu2023visual,zhang2023llavar}, such as Q-Former\cite{li2023blip2}. Altogether, these components accumulate to models with billions of parameters, necessitating a significant amount of memory, computational resources, and training data to achieve their superior performance in VDU tasks.

\section{\AlgoNameNoSpace}
\label{sec:method}

We suggest a new OCR-free document understanding method called \AlgoNameNoSpace.
Our approach revolves around the need to enable interaction between the visual features and the language prompt. To do so, new layers named \Vilma are incorporated into the vision encoder architecture as described in \cref{sec:arch}. During these interactions, a specifically designed pre-training stage (LMPM) guides the vision encoder to concentrate on the pertinent text patches within the document image in relation to the prompt(\cref{sec:pretrain}).

\subsection{Architecture Enhancements Enable Early Prompt Interactions} 
\label{sec:arch}
OCR-free architectures are typically composed of two main components: a visual encoder $\mathcal{M}_V$ responsible for encoding an input document image $X$ into visual features $\hat{Z}$:
\begin{equation}
\hat{Z} = \mathcal{M}_{\scaleto{V}{3.5pt}}\left(X\right)
\end{equation}
and a language model $\mathcal{M}_L$, that receives both the encoded image and the user prompt $\mathbf{p}$ as an input to produce the final prediction $\hat{\mathcal{Y}}$:
\begin{equation}
\hat{\mathcal{Y}}= \mathcal{M}_{\scaleto{L}{3.5pt}} \left(\mathbf{p}, \hat{Z}\right)
\end{equation}
Our objective is to improve the performance over document understanding tasks by introducing the prompt sequence earlier in the model. 
To this end, instead of having a visual representation of the input document independent of the prompt, \AlgoNameNoSpace's encoder $\mathcal{M}_{\scaleto{V}{3.5pt}}^p$ encodes the document image with respect to the language prompt (a question, instruction, etc.) to produce prompt-aware features $\hat{Z}_\mathbf{p}$: 
\begin{equation}
    \hat{Z}_\mathbf{p} = \mathcal{M}_{\scaleto{V}{3.5pt}}^p \left(\mathbf{p}, X\right)
\end{equation}
When integrated with our proposed pre-training scheme which we present in \cref{sec:pretrain}, this enables the encoder to focus on more relevant text patches of the image with respect to the prompt, as demonstrated at the right side of Fig.~\ref{fig:teaser}.

In both previous approaches and in ours, the prompt is also used as an input to the language model, such that the overall model $\mathcal{M}$ formulation reads:
\begin{equation}
    \mathcal{M}\left(\mathbf{p}, X\right) 
    = \mathcal{M}_{\scaleto{L}{3.5pt}} (\mathbf{p}, \hat{Z}_\mathbf{p})
    =\mathcal{M}_{\scaleto{L}{3.5pt}} \left(\mathbf{p}, \mathcal{M}_{\scaleto{V}{3.5pt}}^p \left(\mathbf{p}, X\right)\right)
\end{equation}
Next we specify the architectural enhancements in $\mathcal{M}_{\scaleto{V}{3.5pt}}^p$, to enable interaction between visual feature maps and the prompt \textbf{p}.

\vspace{-3mm}
\subsubsection{Vision-Language Merging Attention} \label{sec:vilma}
\begin{figure}[tb]
  \centering
  \caption{\textbf{An overview of the \AlgoName architecture.} The encoded prompt serves as an input for every ViLMA layer, at the end of each encoding stage (top). The goal of the ViLMA layers is to provide the encoder with prompt guidance during the down-sampling process. The encoded prompt is input through a cross attention layer before down-sampling (bottom).}
    \includegraphics[width=1.0\linewidth]{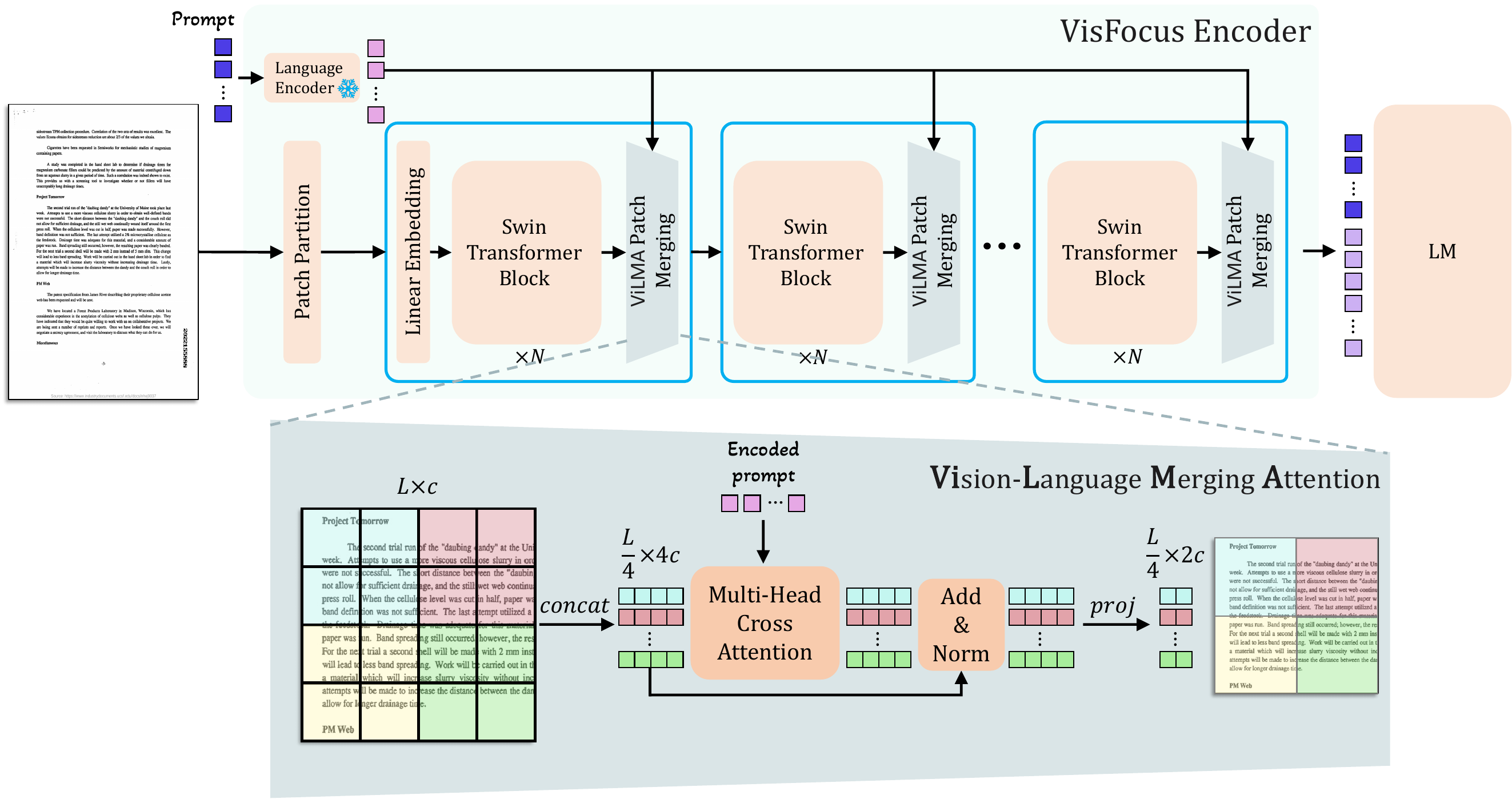}
  \label{fig:arch}
\vspace{-7mm}
\end{figure}

The vision encoder $\mathcal{M}_{\scaleto{V}{3.5pt}}$ of \AlgoName is based on the Swin transformer\cite{liu2022swin}, chosen for its hierarchical architecture designed to capture both local and global information effectively.
The model merges neighbouring patches via patch merging layers, and thus aggregates information into larger more abstract representations at higher scales.
To promote the model's attention towards patches relevant to the input prompt, we chose to intervene at the patch merging layers and modify them accordingly, ensuring that the captured information is contingent on the prompt. We term the newly introduced patch merging mechanism \VilmaNoSpace, which stands for VIsion-Language Merging Attention. The upper part of \cref{fig:arch} illustrates where the \Vilma layers are incorporated in SwinV2 transformer instead of the original patch merging layers, and the mechanism of a single \Vilma layer is shown at the bottom. 

Similarly to the original Swin patch-merging layers, \Vilma concatenates the features of each group of $2\times 2$ neighbouring patches, creating a feature map of size $L/4 \times 4c$ from the original feature map of size $L \times c$, where $L$ is the spatial dimension and $c$ is the corresponding number of channels. 
Then, a linear layer is applied to aggregate the spatial information into higher-level features, concurrently decreasing the feature count by a factor of  $2$, yielding a feature map with dimensions $L/4 \times 2c$.
To ensure that the feature reduction aligns with the prompt, we introduce an interaction layer between the visual features and the prompt. This cross-attention layer is incorporated just before the projection layer, enabling down-sampling to be conducted relative to the prompt.
Following \cite{wang2023visionllm, ganz2023towards} rather than utilizing the original prompt $\textbf{p}$ to guide the encoding of visual features, we employ a frozen language encoder to generate a context-aware representation of the prompt: $\text{emb}(\textbf{p})$. Subsequently, the visual feature map, $\hat{F}$, is passed through a Multi-Head Cross Attention (MHCA) layer \cite{vaswani2017attention}, together with the prompt encoding. This is followed by normalization and additive layers.
Thus, for every feature map:
\begin{equation}
    \tilde{F} = \hat{F} + \text{Norm}\left( \text{MHCA}\left(\hat{F}, \text{emb}(\textbf{p})\right)\right)
\end{equation}
where in the cross attention layer, the visual feature map $\hat{F}$ is used as the query and the prompt embeddings $\text{emb}(\textbf{p})$ are used as both the keys and values.

\subsection{Pre-training Scheme for Prompt-Aware Focusing} \label{sec:pretrain}

\begin{figure}[tb]
  \centering
  \caption{\textbf{Training Scheme.} Previous methods only trained the model to read by predicting the OCR of the document (Stage I). We suggest an addition Localized Masked Prompt Modeling (Stage II) step to train the model to focus on a specific area of text inside the document.}
  \includegraphics[width=\linewidth]{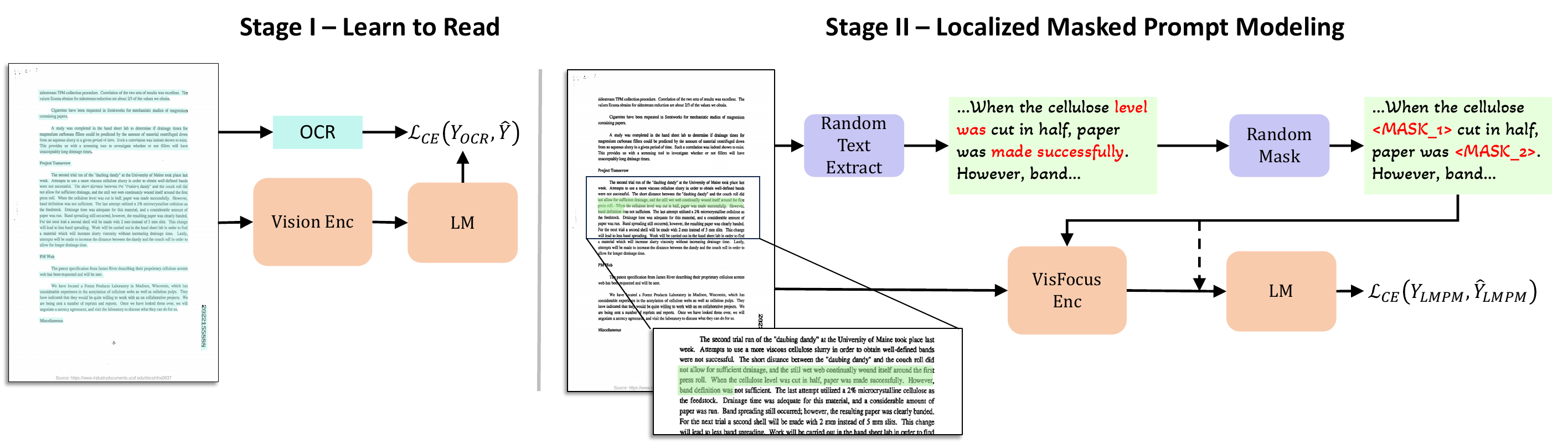}
  \label{fig:train_stages}
\vspace{-7.5mm}
\end{figure}

\begin{figure}[tb]
  \centering
  \caption{\textbf{Attention maps of the last ViLMA layer.} Textual regions relevant to the question tokens are highly activated when performing LMPM pre-training (top) compared to not performing this training stage (bottom). It can be seen that the model focuses its attention not only on the specific input word but also on related words, e.g when performing cross attention with the word ``diameter'' it focuses on the words ``under-ream'' and ``180 degrees''.}
  \includegraphics[width=0.98\linewidth]{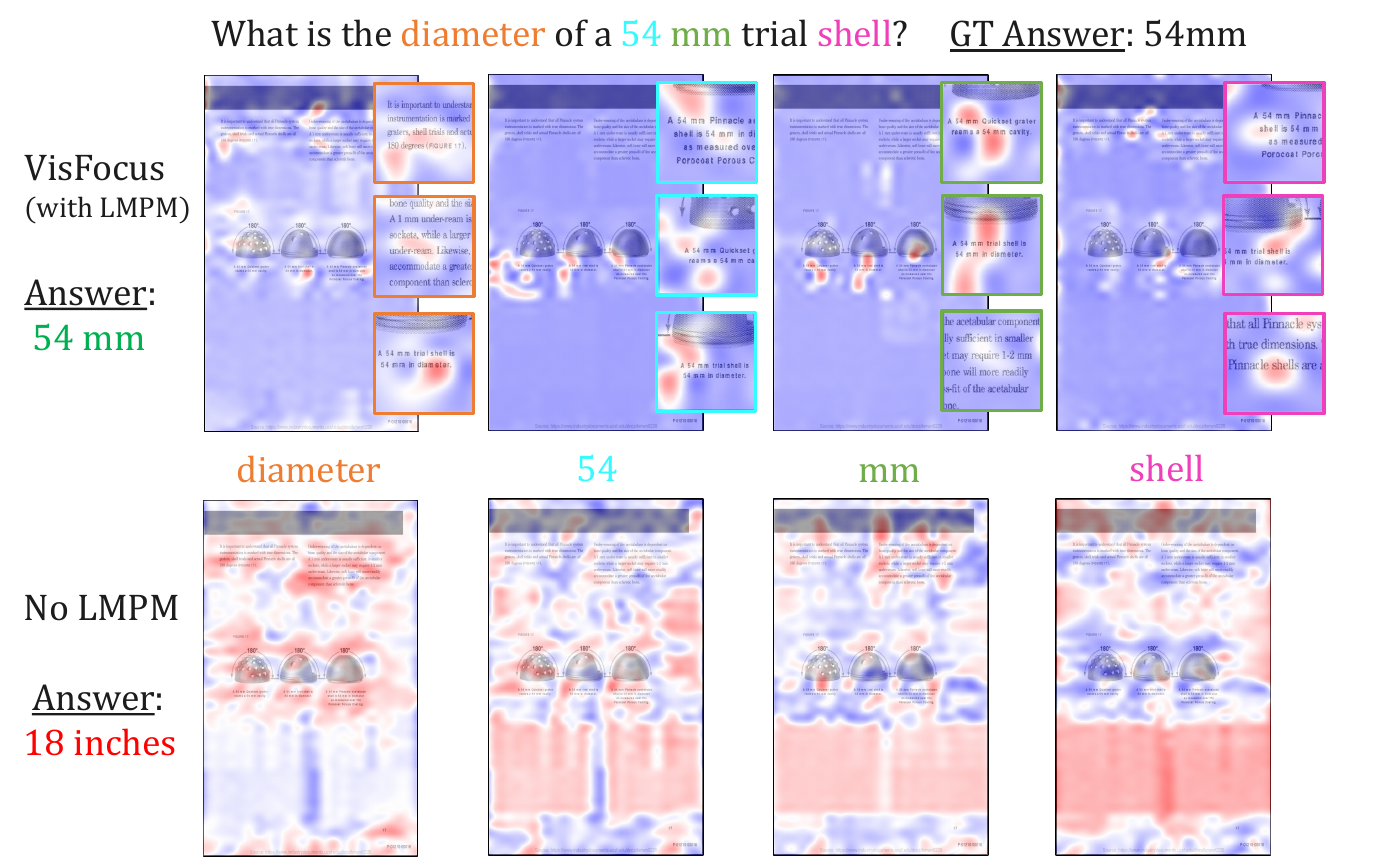}
  \label{fig:attn_maps}
\end{figure}
While incorporating \Vilma facilitates the model's interaction with the user prompt during downsampling, it does not explicitly guide the vision encoder towards focusing on the most relevant text patches. To address this, we introduce a new pre-training task called \textit{LMPM} (Localized Masked Prompt Modeling), as an additional stage in the pre-training scheme outlined in \cref{fig:train_stages}. The overall training process comprises three stages: (1) an LtR (\textbf{L}earn \textbf{t}o \textbf{R}ead) \cite{kim2021donut, lee2023pix2struct} stage, (2) an LMPM stage and (3) a fine-tuning stage over the downstream task.

\subsubsection{Learn to Read (LtR)} 
The objective of this stage is to equip the model with the ability to comprehend text effectively. The crucial role of this step has been demonstrated in OCR-Free literature \cite{kim2021donut, lee2023pix2struct},where its significance in improving the model's ability to process textual information within documents has been shown. Hence, for this basic stage, we align with previous work and adopt the pre-training task of predicting the words in the document in their order of appearance, supervised by external annotations or OCR transcription of the text. The corresponding loss function reads:
\begin{equation}
\label{eq:LtR}
    \mathcal{L}_{\scaleto{LtR}{3.5pt}} = \mathcal{L}_{\scaleto{CE}{3.5pt}}\left(\mathcal{M}(X),Y_{\scaleto{OCR}{3.5pt}}\right)
\end{equation}
where $\mathcal{L}_{CE}$ denotes the cross entropy loss, and $Y_{\scaleto{OCR}{3.5pt}}$ is the top-to-bottom and left-to-right raster-scan order of the text in the input document $X$.

\subsubsection{Localized Masked Prompt Modeling (LMPM)} The primary objective of this stage is to guide the model's attention towards the pertinent sections of the document. To address this, we leverage T5's \cite{raffel2020exploring} denoising objective: for a given sequence, a portion of the tokens are randomly masked and replaced by sentinel tokens, where spans of consecutive tokens are assigned to a single sentinel token. The task is to predict the omitted spans of tokens, separated by the same sentinel tokens utilized in the input sequence.

While adopting this general approach, instead of processing the entire document text, we opt to randomly sample a local span of tokens as illustrated on the right side of \cref{fig:train_stages}. We then apply the Masked Language Modeling (MLM) task to this span while keeping the visual text visible. This masked span $\mathbf{s}$ is subsequently used as a prompt, as we apply the cross entropy loss to predict the masked tokens:
\begin{equation}
    \mathcal{L}_{\scaleto{LMPM}{3.5pt}} = \mathcal{L}_{\scaleto{CE}{3.5pt}}\left(\mathcal{M}(\mathbf{s}, X),Y_{\scaleto{LMPM}{3.5pt}}\right)
\end{equation}
where $Y_{\scaleto{LMPM}{3.5pt}}$ is the set of masked tokens, and $\mathbf{s}$ is the masked sampled sequence.
Given that the LtR task remains constant and is not dependent on the prompt, this stage is trained with the original Swin patch merging layers. The integration of \Vilma layers into the architecture occurs only in the LMPM stage where those layers are randomly initialized.

While feeding the prompt embedding directly to the language model is the common practice, and has been shown to be effective \cite{liu2023improved,li2023blip2,bai2023qwenvl, alayrac2022flamingo,chen2023pali3}, 
incorporating the prompt as input for both the vision encoder and the language model presents a challenge: the language model might compensate for any missing visual information from the vision encoder, effectively performing the Masked Language Modeling (MLM) task. To address this concern and encourage the vision encoder to develop its own focusing capabilities independently of the language model, both components need to be trained accordingly. 
To foster this independence, inspired by the Dropout technique \cite{srivastava2014dropout}, we adopt a strategy where the prompt is concatenated to the language model's input with a certain probability $\rho\in[0,1]$ and omitted otherwise, such that,
\begin{equation}
\label{eq:concat-dropout}
    \hat{\mathcal{Y}} = 
    \begin{cases}
        \mathcal{M}_{\scaleto{L}{3.5pt}}\left(\mathbf{p}, \hat{Z}\right)\hspace{0.5cm} \text{if } \epsilon < \rho\\
        \mathcal{M}_{\scaleto{L}{3.5pt}}\left(\hat{Z}\right)\hspace{.9cm} otherwise
    \end{cases}
\end{equation}
where $\epsilon\in\left[0,1\right]$ is sampled uniformly at random at every training step. 
The benefits of applying LMPM are illustrated in \cref{fig:attn_maps} where the attention maps of the last \Vilma layer are depicted relative to words from the input query. In the top panel, where LMPM is utilized, the model exhibits a focused attention on relevant words, while in the bottom panel, where the LMPM stage is omitted, the model's attention appears scattered.  
For instance, when examining attention relative to the word ``diameter'' the model trained with LMPM focuses on related terms such as ``under-ream'' and ``180 degrees'' showcasing its improved ability to discern contextually relevant information.
Further quantitative analysis of LMPM and \cref{eq:concat-dropout} are demonstrated in \cref{ssec:ablation_study}.

\section{Experiments} \label{sec:experiments}
\begin{figure}[tb]
  \centering
  \caption{\textbf{Qualitative comparison.} We present examples from the DocVQA validation set, demonstrating our model's ability to accurately answer questions on denser documents compared to the baseline.}
  \includegraphics[width=1\linewidth]{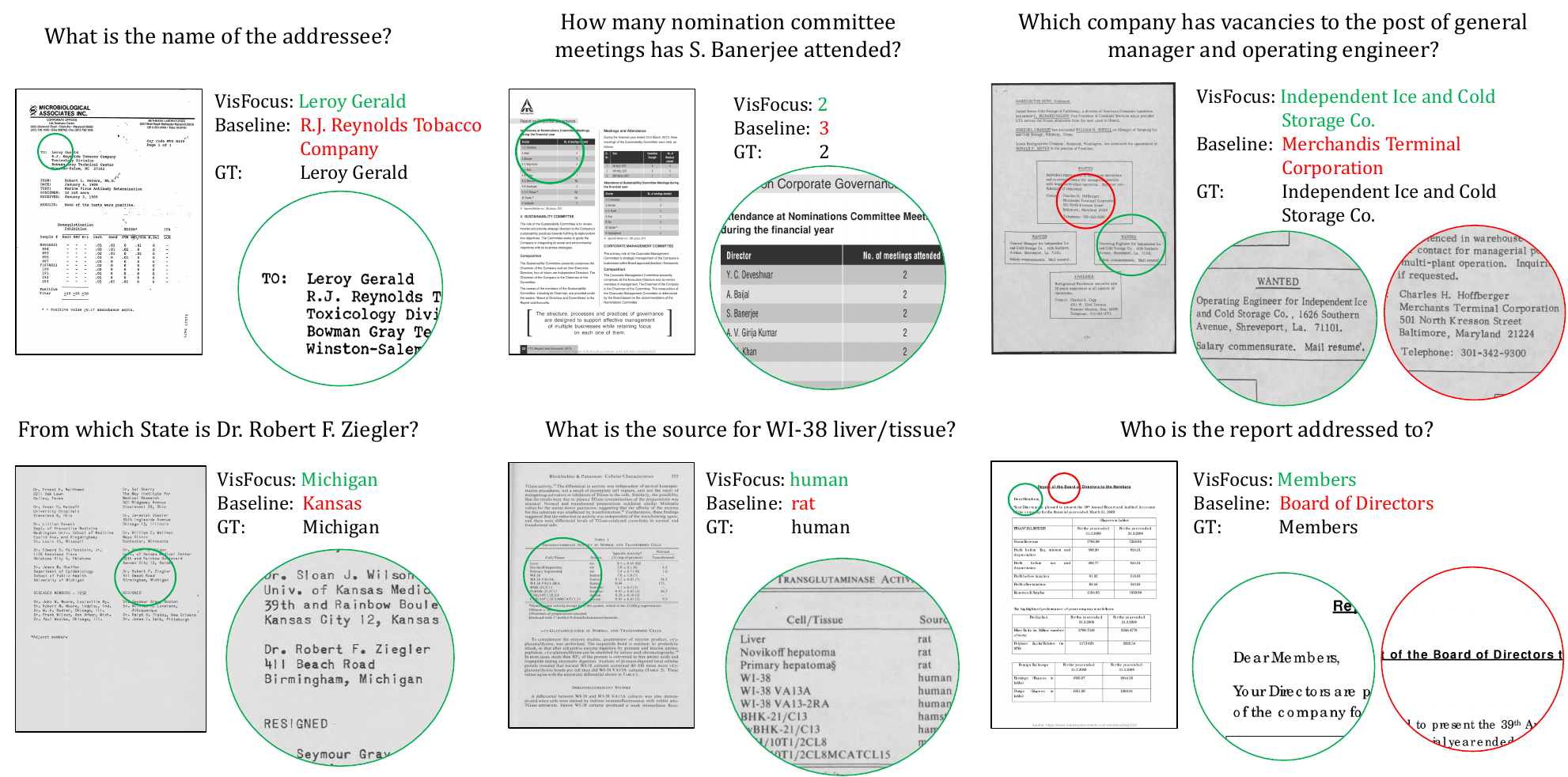}
  \label{fig:bl_comparison}
\vspace{-6mm}
\end{figure}

We first present our experimental setup in \cref{ssec:experimental_setup}, followed by results comparing to previous approaches on various Document VQA benchmarks in \cref{ssec:comparison_previous}. We then present an ablation study showing the contribution individual components of our method and their synergy in \cref{ssec:ablation_study}. 
Finally we show a performance analysis \AlgoName on varying document densities, showcasing its even increasing contribution for dense documents. Implementation details and further information on all datasets can be found in the appendix.

\subsection{Experimental Setup} \label{ssec:experimental_setup}
As stated in \cref{sec:method}, \AlgoName is composed of three main components: a vision encoder trained to extract visual features from high-resolution documents images, a projection module, and an Language Model (LM) that receives both the prompt and the projected visual features, yielding the final output.

\subsubsection{Implementation Details}
\AlgoName utilizes SwinV2\cite{liu2022swin} and T5 models \cite{2019t5} as the vision encoder and language model, pre-trained on ImageNet-1K~\cite{deng2009imagenet} and ``Colossal Clean Crawled Corpus'' (C4)~\cite{raffel2020exploring}, respectively. In all our experiments, we utilize the SwinV2-Small variant. 
To align the output visual features with the LM, we employ a small multi-layer perceptron. 
This module, trained from scratch, projects the vision encoder's output into a shared latent space with the LM's input, before feeding it into the LM as input embeddings.
We introduce two variants of our model: \AlgoNameNoSpace-S and \AlgoNameNoSpace-B, which incorporate the T5-Small and T5-Base variants, respectively.
We compare \AlgoName against corresponding baselines without the ViLMA layers and LMPM pre-training stage. We refer to those as ``Baseline-\{S,B\}''. In all our experiments, for both training and fine-tuning, we train on 8 A100 GPUs with bfloat-precision. We use AdamW \cite{DBLP:journals/corr/abs-1711-05101} optimizer with cosine annealing \cite{DBLP:journals/corr/LoshchilovH16a} learning rate scheduler and warm-up. We train on high resolution input images of $1536\times 768$.  The complete implementation details and training recipes can be found in the Appendix.

\vspace{-2mm}
\subsubsection{Datasets and Metrics}
At the pre-training stages, we train our models on the IDL dataset \cite{biten2022ocr} of document pages with OCR annotations. 
We evaluate our method against previous approaches over five different VQA benchmarks containing various domains: documents, infographics, charts and book covers.
DocVQA\cite{mathew2021docvqa}, a subset of IDL, consists of $14k$ document images and $40k$ questions. 
InfographicsVQA (InfoVQA)\cite{mathew2022infographicvqa} consists of $5k$ infographic images crawled from the web and $30k$ questions. 
ChartQA\cite{masry-etal-2022-chartqa} contains chart images with questions requiring both visual and logical reasoning. It consists of $9.6K$ human-written questions and $23.1K$ generated questions based on summaries.
OCR-VQA\cite{mishraICDAR19} is a large-scale dataset of $200k$ book cover images and $1M$ questions.
AI2 Diagrams (AI2D)\cite{Kembhavi2016ADI} consists of $5K$ grade-school science diagrams, with corresponding multiple-choice questions.
For DocVQA and InfoVQA we report Average Normalized Levenshtein Similarity (ANLS) metric \cite{mathew2021docvqa}, for ChartQA we follow \cite{masry-etal-2022-chartqa} and report average Relaxed Accuracy (RA), and on OCR-VQA and AI2D we report Exact Match (EM). All of which are defined in the appendix for brevity. More information about datasets and metrics can found in Appendix.

\vspace{-3mm}
\subsubsection{Baselines}
\label{sec:baselines}
We compare \AlgoName to state-of-the-art OCR-free approaches for the small and base model size category. In the small category we compare to Dessurt \cite{davis2022dessurt} and Donut \cite{kim2021donut} which are pre-trained using the LtR task (\cref{eq:LtR}) on a corpus of real and synthetic documents. In the base category we compare to Pix2Struct-B \cite{lee2023pix2struct} which uses screen-shot parsing as a pre-training task, and ScreenAI \cite{baechler2024screenai} which uses screen user interfaces for pre-training. For completeness, we present the results of other notable LVLMs.

\vspace{-2mm}
\subsection{Comparison to Previous Approaches} \label{ssec:comparison_previous}
\vspace{-6mm}
\setlength{\tabcolsep}{4pt}
\begin{table}
    \begin{center}
        \caption{\textbf{Comparison with previous OCR-Free methods on VQA benchmarks.} \model outperforms previous methods of comparable scale, even when trained on substantially less pre-training data. We report ANLS on DocVQA and InfoVQA, Relaxed Accuracy (RA) on ChartQA and Exact Match (EM) on OCR-VQA and AI2D. In fully-trained methods, we only state total number of parameters.}
        \vspace{-3mm}
        \label{table:main}
        \resizebox{\textwidth}{!}{
        \begin{tabular}{cl|c|cccccc}
            \toprule
            & \multirow{2}{*}{Method} & \#params & \multirow{2}{*}{DocVQA} & \multirow{2}{*}{InfoVQA} & \multirow{2}{*}{ChartQA} & \multirow{2}{*}{OCR-VQA} & \multirow{2}{*}{AI2D}\\
            & & \small{(Trainable / Total)} & \tiny{ANLS} & \tiny{ANLS} & \tiny{RA} & \tiny{EM} & \tiny{EM}\\
            \hline & \\[-2ex]
            \multirow{4}{*}{\rotatebox{90}{{Large}}}
            & \textcolor{dGray}{UReader\cite{ye2023ureader}}  & \textcolor{dGray}{86M / 7B }  & \textcolor{dGray}{65.4}   & \textcolor{dGray}{42.2}   & \textcolor{dGray}{59.3}  & \textcolor{dGray}{-   }  & \textcolor{dGray}{-   }     \\
            & \textcolor{dGray}{Pix2Struct-L\cite{lee2023pix2struct}}  & \textcolor{dGray}{1.3B}       & \textcolor{dGray}{76.6}   & \textcolor{dGray}{40.0}   & \textcolor{dGray}{58.6}  & \textcolor{dGray}{\textbf{71.3}}  & \textcolor{dGray}{42.1}     \\
            & \textcolor{dGray}{mPlugDocOwl\cite{ye2023mplug}}  & \textcolor{dGray}{7B  }       & \textcolor{dGray}{62.2}   & \textcolor{dGray}{-   }   & \textcolor{dGray}{57.4}  & \textcolor{dGray}{-   }  & \textcolor{dGray}{-   }     \\
            & \textcolor{dGray}{PaLi-3\cite{chen2023pali3}}  & \textcolor{dGray}{5B  }       & \textcolor{dGray}{\textbf{87.6}}   & \textcolor{dGray}{\textbf{57.8}}   & \textcolor{dGray}{\textbf{76.7}}  & \textcolor{dGray}{70.0}  & \textcolor{dGray}{\textbf{75.2}   }     \\
            \cline{2-8} & \\ [-2ex]
            \multirow{4}{*}{\rotatebox{90}{{Small}}}
            & Dessurt\cite{davis2022dessurt} & 127M & 63.2 & - & - & - & - \\
            & Donut\cite{kim2021donut} & 176M & 67.5 & 11.6 & 41.8 & 66.0 & - \\
            & \cellcolor{gray!20}Baseline-S & \cellcolor{gray!20}110M & \cellcolor{gray!20}67.0 & \cellcolor{gray!20}24.7 & \cellcolor{gray!20}49.3 & \cellcolor{gray!20}66.6 & \cellcolor{gray!20}\textbf{42.7} \\
            & \cellcolor{gray!20}\AlgoNameNoSpace-S & \cellcolor{gray!20}132M / 171M & \cellcolor{gray!20}\textbf{68.6}\gcol{\texttt{+}1.6} & \cellcolor{gray!20}\textbf{28.5}\gcol{\texttt{+}3.8} & \cellcolor{gray!20}\textbf{53.0}\gcol{\texttt{+}3.7} & \cellcolor{gray!20}\textbf{67.3}\gcol{\texttt{+}0.7} & \cellcolor{gray!20}42.6\rcol{\texttt{-}0.1} \\
            \cline{2-8} & \\[-2ex]
            \multirow{4}{*}{\cellcolor{white}\rotatebox{90}{{Base}}}
            & ScreenAI-B\cite{baechler2024screenai} & 670M & 50.7 & 19.6 & 54.0 & 54.8 & - \\
            & Pix2Struct-B\cite{lee2023pix2struct} & 282M & 72.1 & \textbf{38.2} & 56.0 & 69.4 & 40.9 \\
            & \cellcolor{gray!20}Baseline-B & \cellcolor{gray!20}273M & \cellcolor{gray!20}71.7 & \cellcolor{gray!20}26.8 & \cellcolor{gray!20}52.5 & \cellcolor{gray!20}66.9 & \cellcolor{gray!20}45.6 \\
            & \cellcolor{gray!20}\AlgoNameNoSpace-B & \cellcolor{gray!20}295M / 408M & \cellcolor{gray!20}\textbf{72.9}\gcol{\texttt{+}1.2} & \cellcolor{gray!20}31.9\gcol{\texttt{+}5.1} & \cellcolor{gray!20}\textbf{57.1}\gcol{\texttt{+}4.6} & \cellcolor{gray!20}\textbf{70.0}\gcol{\texttt{+}3.1} & \cellcolor{gray!20}\textbf{47.8}\gcol{\texttt{+}2.2} \\
            \bottomrule
            \end{tabular}
            }
        \end{center}
    \end{table}
  \vspace{-6mm}
\setlength{\tabcolsep}{1.4pt}

We compare the performance of our model to previous methods over five different benchmarks in \cref{table:main} as specified in \cref{sec:baselines}.
The listed methods are categorized into three groups by model size. Those of base and small sizes are compared against our \AlgoName variants of the same size category.
Large models with billions of parameters are included for completeness, as those require substantially more data and computational resources.
\AlgoName improves over the baseline across all datasets in the base category and most datasets in the small category. It can be seen that \AlgoNameNoSpace-S and \AlgoNameNoSpace-B yield a performance gap of $\mathbf{+1.6}$, $\mathbf{+1.2}$ points on DocVQA, $\mathbf{+3.8}$, $\mathbf{+5.1}$ on InfoVQA, $\mathbf{+3.7}$, $\mathbf{+4.6}$ on ChartQA, $\mathbf{+0.7}$, $\mathbf{+3.1}$ on OCR-VQA and $\textbf{-0.1}$, $\mathbf{+2.2}$ on AI2D over their corresponding baselines.
Notice that the additional parameter count attributed to the introduction of ViLMA layers are an order of magnitude smaller than the model size. 
Our approach achieves state-of-the-art performance, surpassing prior methods on four out of five benchmarks for the small category and on all benchmarks for the base category.
While the performance of \AlgoName over InfoVQA is significantly better than other methods oriented at equipping the model with reading capabilities (e.g., Donut and Dessurt), it is still lower than Pix2Struct. Considering that reasoning about infographics not only requires reading textual information but also processing other visuals, the gap is likely attributed to our focus on reading the most relevant parts of the document compared to Pix2Struct, which trains with more diverse pre-training tasks and over a larger diverse dataset, not publicly available.

\vspace{-3mm}
\subsection{Ablation Study and Empirical Analysis} \label{ssec:ablation_study}
\vspace{-7mm}

\setlength{\tabcolsep}{4pt}
\begin{table}[]
\begin{center}
\caption{\textbf{Breaking down the contributions of VisFocus' main components}. ANLS for DocVQA and RA for ChartQA are reported.}
\label{table:ablations}
\vspace{-2mm}
\resizebox{\textwidth}{!}{
\begin{tabular}{l|cccc|cccc|ccc}
\toprule
\multirow{2}{*}{Method} & & \multicolumn{2}{c}{{Prompt Interaction}}  & & & \multicolumn{2}{c}{{LMPM Stage}} & & & DocVQA & ChartQA \\
& &\scriptsize{Concat} & \scriptsize{\Vilma} & & & \scriptsize{Concat} & \scriptsize{Alternate} & & & \tiny{ANLS}   & \tiny{RA} \\
\hline & \\[-2ex]

Baseline-B                      & & \cmark &  & & &  & & &  & 70.9 & 52.5 \\
\hspace{2mm} $+$\Vilma          & & \cmark & \cmark & & & & &  &  & 71.3 & 54.7 \\
\hspace{4mm} $+$LMPM            & & \cmark & \cmark & & & \cmark & & &  & 71.8 & 55.7 \\

\rowcolor{Gray}\hspace{6mm} $+$concat (\cref{eq:concat-dropout})& & & & & & & & & & &\\
\rowcolor{Gray}\hspace{8mm} (\AlgoNameNoSpace-B) & & \multirow{-2}{*}{\cmark} & \multirow{-2}{*}{\cmark} & & & \multirow{-2}{*}{\cmark} & \multirow{-2}{*}{\cmark} & & & \multirow{-2}{*}{\textbf{72.2}} & \multirow{-2}{*}{\textbf{57.1}} \\

\bottomrule
\end{tabular}}
\end{center}
  \vspace{-7mm}
\end{table}
\setlength{\tabcolsep}{1.4pt}

We conduct an ablation study breaking down the impact of each component of \AlgoName individually and pilling those up gradually to showcase the synergy when using both architectural enhancements and the pre-training scheme together. We evaluate \AlgoNameNoSpace-B and report ANLS on the formal validation set of DocVQA for simplicity and the formal test set of ChartQA in \cref{table:ablations}. Each row in the table represents each of our contributions added independently, starting from our baseline, with all of the examined components disabled.

\vspace{-4mm}
\subsubsection{\VilmaNoSpace} We first quantify the contribution of architectural enhancements alone. As can be seen in the second line of \cref{table:ablations}, the transition from Swin's patch-merging layers to \Vilma layers add $\mathbf{+0.4}$ and $\mathbf{+2.2}$ points on DocVQA and ChartQA respectively.

\vspace{-3mm}
\subsubsection{LMPM} 
The contribution of the \Vilma layers fulfills its potential when complemented by an appropriate pre-training task to encourage the encoded visual features to focus on relevant text patches. This is reflected by the additional $\textbf{+0.5}$ and $\textbf{+1.0}$ points on DocVQA and ChartQA respectively, as specified in the third row of \cref{table:ablations}.
To ensure that the vision encoder attends to the prompt (provided via the ViLMA layers) and does not ignore it, we employ \cref{eq:concat-dropout} to randomly skip the concatenation of the prompt to the LM’s input. 
The final row in \cref{table:ablations} quantifies the benefits of this technique by $\mathbf{+0.4}$ and $\mathbf{+1.4}$ point gains on DocVQA and ChartQA, respectively.

\subsubsection{Prompt Insertion Methods} 
\begin{figure}[ht]
  \centering
  \begin{minipage}[t]{0.45\textwidth}
    \centering
    \captionof{table}{\textbf{Prompt Insertion Methods.} Inserting the prompt via \Vilma layers improves results compared to previous approaches with only LtR pre-training applied (without LMPM). ``Render''=question is rendered on the document image.}
    \label{table:prompt_insert}
            \resizebox{\textwidth}{!}{
        \begin{tabular}{lcc}
            \toprule
            Injection & DocVQA & ChartQA \\
            Strategy & \tiny{ANLS} & \tiny{RA}   \\
            \hline & \\[-2ex]
            Baseline (LM-only)  & 70.9 & 52.5 \\
            Render (Pix2Struct)  & 70.6 & 52.2 \\
            \rowcolor{Gray}\AlgoName(ViLMA) & \textbf{71.3} & \textbf{54.7} \\
            \bottomrule
        \end{tabular}
        }

  \end{minipage}
  \hspace{1em}
  \begin{minipage}[t]{0.5\textwidth}
    \captionof{table}{\textbf{\Vilma Layers Locations.} Using \Vilma layers instead of all of the patch merging layers improves results compared to replacing part of the layers.}
    \label{Tables:vilma_abl}
            \resizebox{\textwidth}{!}{
        \begin{tabular}{lccc}
            \toprule
            & Integration & DocVQA & ChartQA \\
            & Stages & \tiny{ANLS} & \tiny{RA}   \\
            \hline & \\[-2ex]
            Baseline & \texttt{none} & 70.9 & 52.5 \\
            \hline & \\[-2ex]
            VF-Early & \texttt{[1,2]}   & 71.0 & 54.1 \\
            VF-Mid & \texttt{[2,3]}  & 71.3 & 54.4 \\
            VF-Late & \texttt{[3,4]}  & 71.6 & 55.3 \\
            \rowcolor{Gray}VF-All & \texttt{[1,2,3,4]} & \textbf{72.2} & \textbf{57.1} \\
            \bottomrule
        \end{tabular}

        }

  \end{minipage}
  \vspace{-7mm}
\end{figure}

In this section we examine alternative ways to insert the prompt to the model, comparing our \Vilma layers to previous approaches. 
\cref{table:prompt_insert} compares between (1) the baseline approach of inserting the prompt to the language model alone (as done in e.g. Donut and Dessurt), (2) the approach suggested in Pix2Struct of rendering the prompt on top of the input image, and (3) our newly introduced \Vilma layers. \Vilma layers insert the prompt directly to the vision encoder patch-merging layers in addition to the LM input. This approach yields an improvement over the baseline, achieving gains $\mathbf{+0.4}$ and $\mathbf{+2.2}$ points on DocVQA and ChartQA, respectively. The rendering approach on contrast, lowers the results compared to the baseline approach. Note that for a fair comparison we pre-trained all of the models in the same way with LtR only (\cref{eq:LtR}) and without LMPM. Hence, we hypothesize that rendering the prompt on top of the image was more effective when applied under the pre-training tasks and data suggested in Pix2Struct, but underperforms when only fine-tuned using this approach.

To further quantify the contribution of substituting patch-merging layers with \Vilma layers within each block, we perform an ablation study as presented in \cref{Tables:vilma_abl}.
Each row of the table corresponds to a fine-tuning experiment where a subset of patch-merging layers are replaced with \Vilma layers. Our findings indicate that integrating \Vilma layers into deeper blocks yields more significant improvements. Specifically, employing \Vilma layers in all blocks results in the highest performance enhancement of $\mathbf{+1.3}$ and $\mathbf{+4.6}$ points, compared to a relatively lower improvement of $\mathbf{+0.7}$ and $\mathbf{+2.8}$ points on DocVQA and ChartQA respectively, when replacing only the last two layers.

\vspace{-3mm}
\subsubsection{Qualitative Analysis} \label{ssec:qualitative}
\cref{fig:bl_comparison} provides qualitative examples comparing Baseline-B and \AlgoNameNoSpace-B. In the Baseline-B, the visual tokens represent the entire document content rather than just what is relevant to the specific prompt. This sometimes leads to incorrect predictions, as the baseline model may extract an answer from irrelevant text patches. 
This effect is visualized in \cref{fig:attn_maps}, where the cross-attention maps inside the \Vilma layers are plotted as heatmaps on top of the input image, showing the interactions between different tokens in the prompt and the visual patches. These visualizations offer clear insights into how \AlgoName learns to subjectively encode the document in relation to the given input query. 

\vspace{-3mm}
\subsubsection{Document Density Analysis} \label{ssec:dda}
\begin{wrapfigure}{r}{0.55\textwidth}
  \centering
    \vspace{-9.5mm}
  \includegraphics[width=0.4\textwidth]{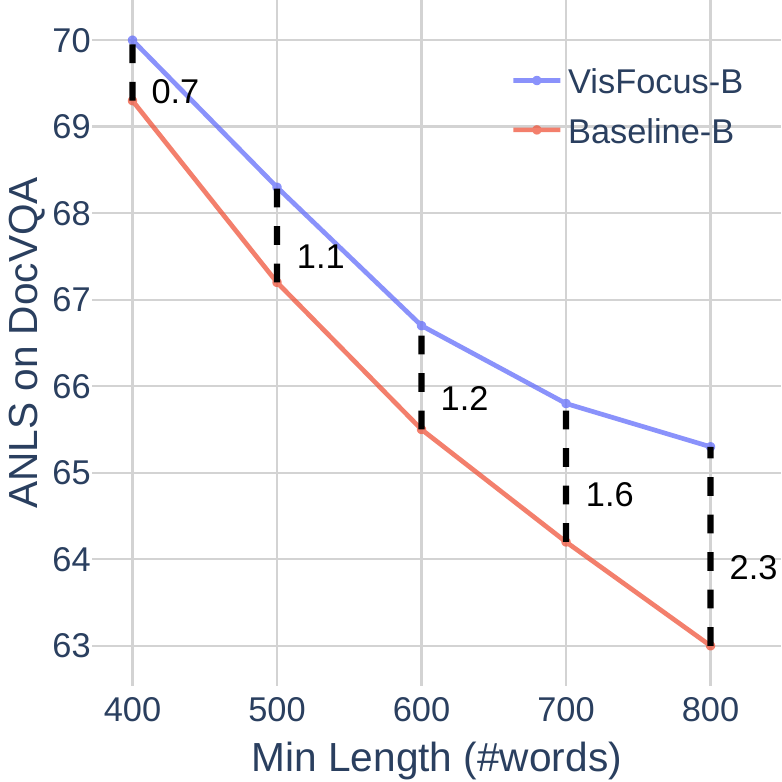}
    \caption{\textbf{Performance vs Number of Words.} The graph shows the ANLS on a subset of the validation set of DocVQA containing at least Min Length words. The marginal gains achieved by \AlgoName increase with the minimum number of words per document. This illustrates the significance of focusing the visual features on specific textual patches for dense documents.}
    \label{fig:anls_vs_num_words}
    \vspace{-6mm}
\end{wrapfigure}
In the following experiment we showcase the benefits of \AlgoNameNoSpace's ability to focus on the most relevant text patches among all, possibly many, irrelevant text patches in dense documents. To this end, we measure the performance on subsets of documents with increasing densities (word counts). This is done by grouping the validation set of DocVQA to overlapping groups according to the minimum number of words ($400$, $500$, $600$, $700$ and $800$). For example, the first group consists of all documents containing at least $500$ words, while the last group consists of all documents with at least  $800$ words. We compare the performance of \AlgoName versus the baseline across these groups in \cref{fig:anls_vs_num_words}. The consistently increasing performance gap on denser documents, ranging from $+0.7$ to $+2.3$ for all documents containing at least $400$ and $800$ words, respectively. This is aligned with our conjecture that in denser documents, focusing on the most relevant text patches to the specific user prompt is even more significant, given the larger amount of redundant information in the document.

\vspace{-2mm}
\section{Conclusions}
\vspace{-2mm}
\label{sec:conclusion}
In this work, we propose a novel way to make the visual encoding in OCR-Free VDU models aware of the user query. Consequently, the model learns to focus on reading the most relevant text in the document. The proposed method, \AlgoNameNoSpace, couples the patch merging layers of a Swin transformer encoder with the user query inside newly introduced Vision-Language Merging Attention (ViLMa) layers. These are trained to focus mostly on encoding text relevant to the user query via a designated Localized Masked Prompt Modeling (LMPM) task. Those complementary components work in synergy to achieve state-of-the-art performance over a variety of document VQA tasks. 

The purpose of this work is to equip the model with prompt-guided reading capabilities, and thus it is encouraged to focus on relevant text. A valid future research direction is the design of additional prompt-aware pre-train tasks, that guide the visual encoder to focus on content relevant to the user query beyond text. Specifically, this has the potential to improve performance on documents containing infographics, charts, and figures as well as on other domains.

\bibliographystyle{splncs04}
\bibliography{egbib}
\newpage

\section*{\centering Supplementary Materials for \\
        VisFocus: Prompt-Guided Vision Encoders for OCR-Free Dense Document Understanding} \label{sec:supplamentry}

\appendix

\blfootnote{*Work done during an internship$\backslash$employment at Amazon}%
\blfootnote{\textsuperscript{\textdagger} Corresponding author: \href{mailto:nivnay@amazon.com}{nivnay@amazon.com}}

\section{\AlgoName Visualization}
\label{apdx: attention maps}

\subsection{LMPM Pre-training}
\begin{figure}[tbh]
  \centering
    \includegraphics[width=1.0\linewidth]{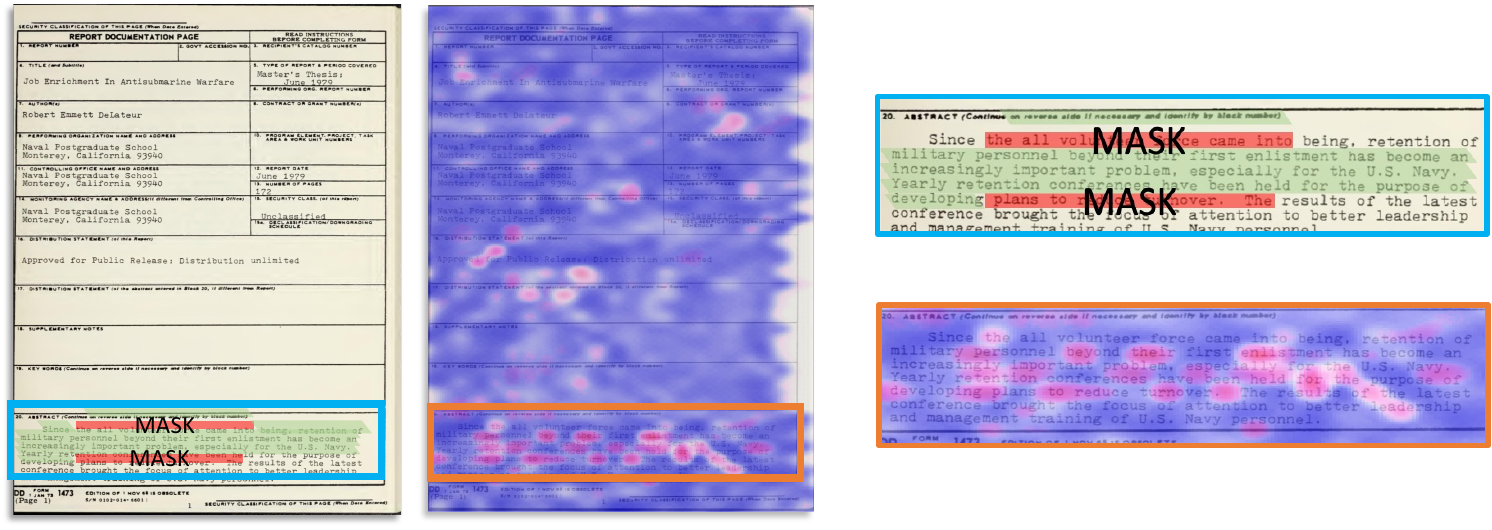}
   \caption{\textbf{Aggregated Token Visualizations for LMPM.} We combine multiple attention maps between textual and visual token from the last layer of our document encoder.  \fcolorbox{lightGreen}{lightGreen}{\textcolor{lightGreen}{-}} denotes randomly sampled text snippets.}
  \label{fig:lmpm_vis}
\end{figure}

To further elucidate the efficacy of LMPM pre-training in focusing where prompt-related textual regions are, we conduct an additional extensive visualization in  \cref{fig:lmpm_vis}, showing multiple text tokens' aggregated attention maps across the document. It can be seen that most of the attention is activated where the sampled text snippet (served as the prompt) originally lies.

\subsection{VQA Fine-tuning}
\begin{figure}[H]
    \centering
    \caption{\textbf{ViLMA Attention Maps.} Attention maps of the last ViLMA layer activated at words from the input prompt, the frames are colored according to the colored prompt tokens and the highly activated visual tokens are explicitly written inside the boxes. Top rows: \AlgoNameNoSpace. Bottom rows: \AlgoName \textbf{without LMPM pre-training}.}
    \begin{subfigure}[H]{1\textwidth}
        \centering
        \includegraphics[width=.8\linewidth]{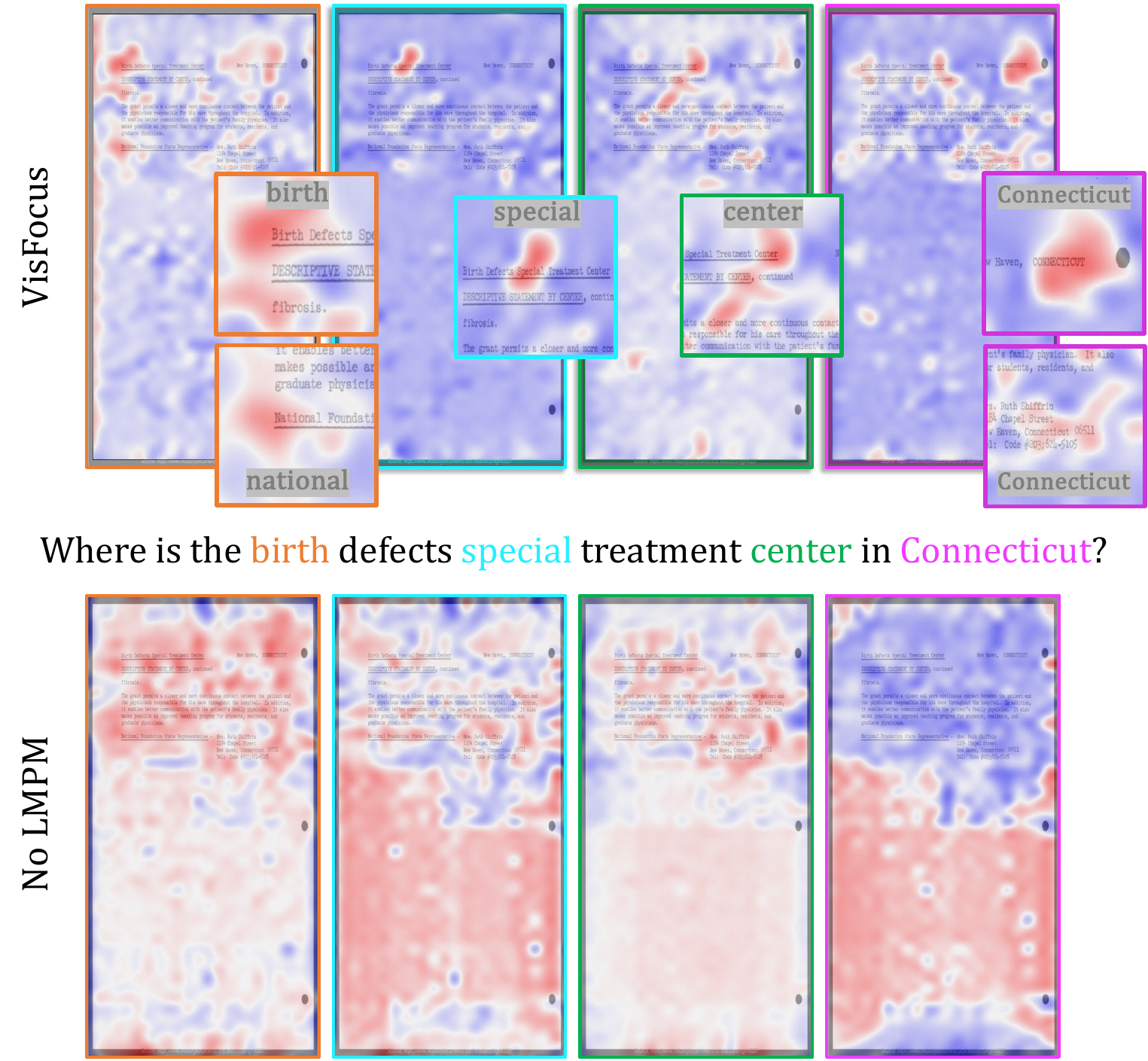}
        \caption{}
        \label{fig:supp_more_vis_1_1}
    \end{subfigure}
  \label{fig:supp_more_vis_1}
  \vspace{-5mm}
\end{figure}

\begin{figure}[ht]
  \centering
    \begin{subfigure}[t]{1\textwidth}
        \centering
        \includegraphics[width=.75\linewidth]{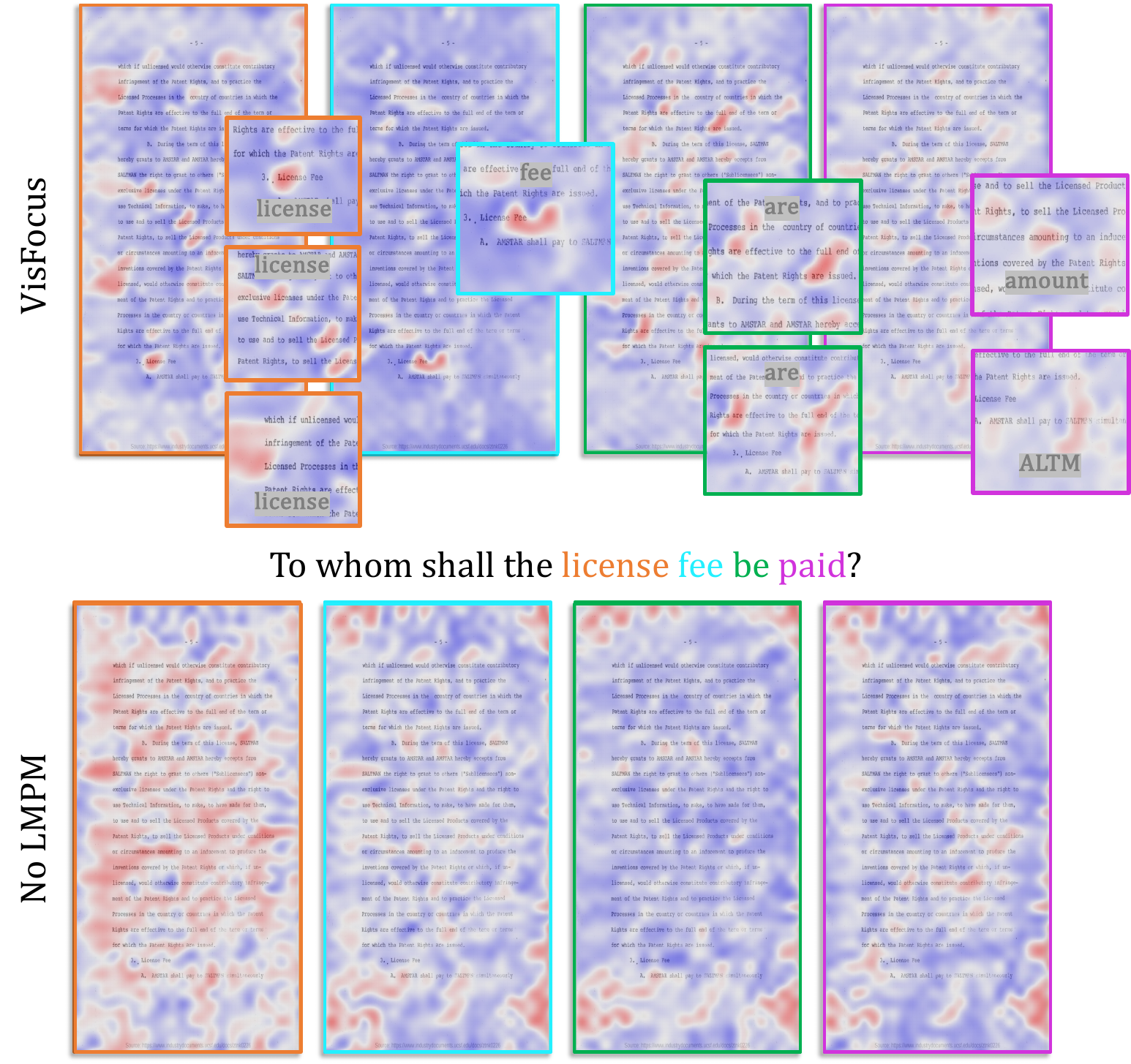}
        \caption{}
        \label{fig:supp_more_vis_1_2}
    \end{subfigure}
    
    \vspace{1pt}
    \rule{\textwidth}{0.5pt}
    \vspace{1pt}

    \begin{subfigure}[t]{1\textwidth}
        \centering
        \includegraphics[width=1\linewidth]{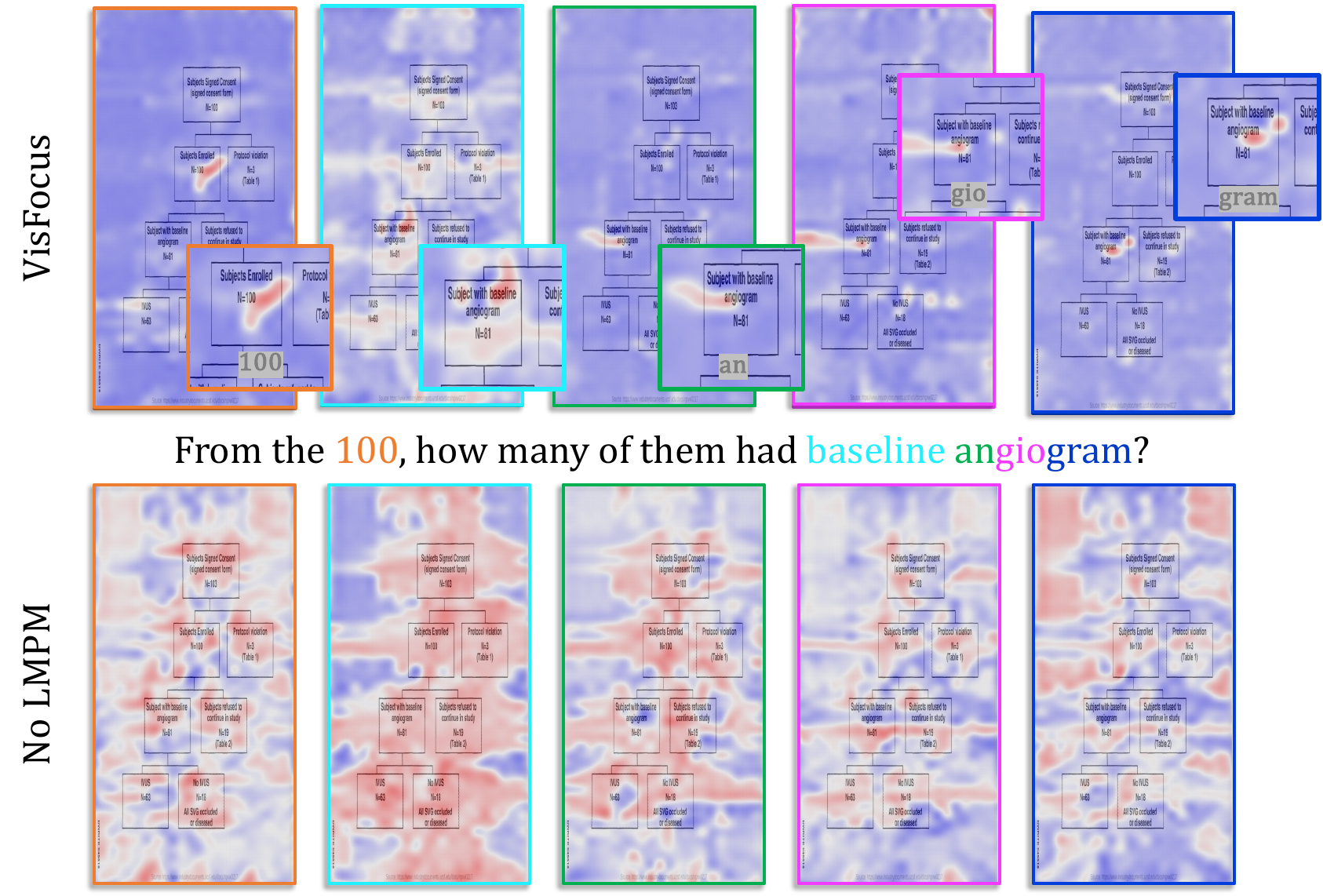}
        \caption{}
        \label{fig:supp_more_vis_1_3}
    \end{subfigure}
    \caption{\cref{fig:supp_more_vis_1} continued.}
    \label{fig:supp_more_vis_2}
\end{figure}

\begin{figure}[ht]
    \centering
    \vspace{-5mm}
    \begin{subfigure}[t]{1\textwidth}
        \centering
        \includegraphics[height=.75\linewidth]{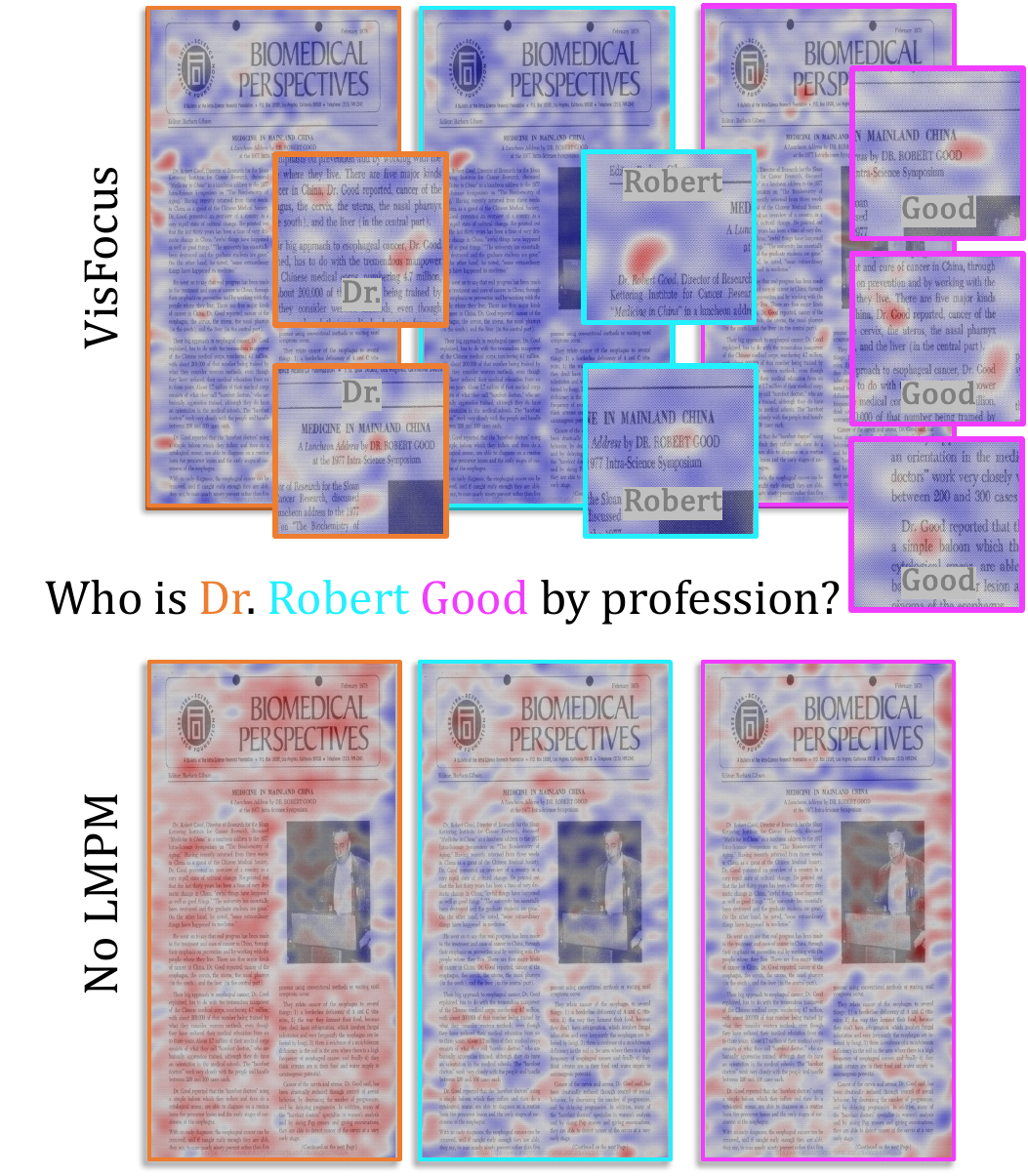}
        \caption{}
        \label{fig:supp_more_vis_1_4}
    \end{subfigure}
    
    \vspace{1pt}
    \rule{\textwidth}{0.5pt}
    \vspace{1pt}
    
    \begin{subfigure}[ht]{.4\textheight}
        \centering
        \includegraphics[width=1\linewidth]{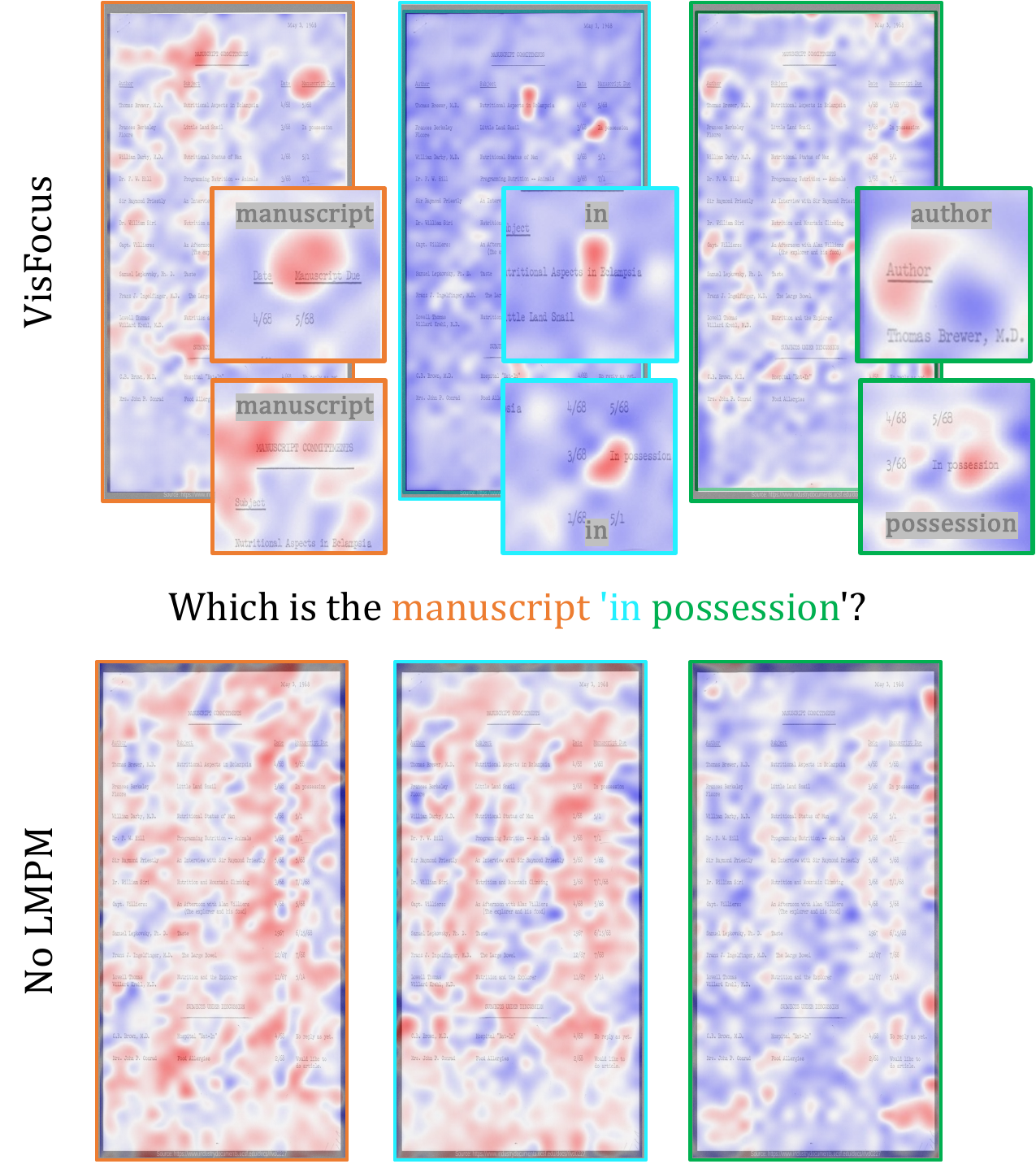}
        \caption{}
        \label{fig:supp_more_vis_3_4}
    \end{subfigure}
    \vrule
    \begin{subfigure}[ht]{.2\textheight}
        \centering
        \includegraphics[width=1\linewidth]{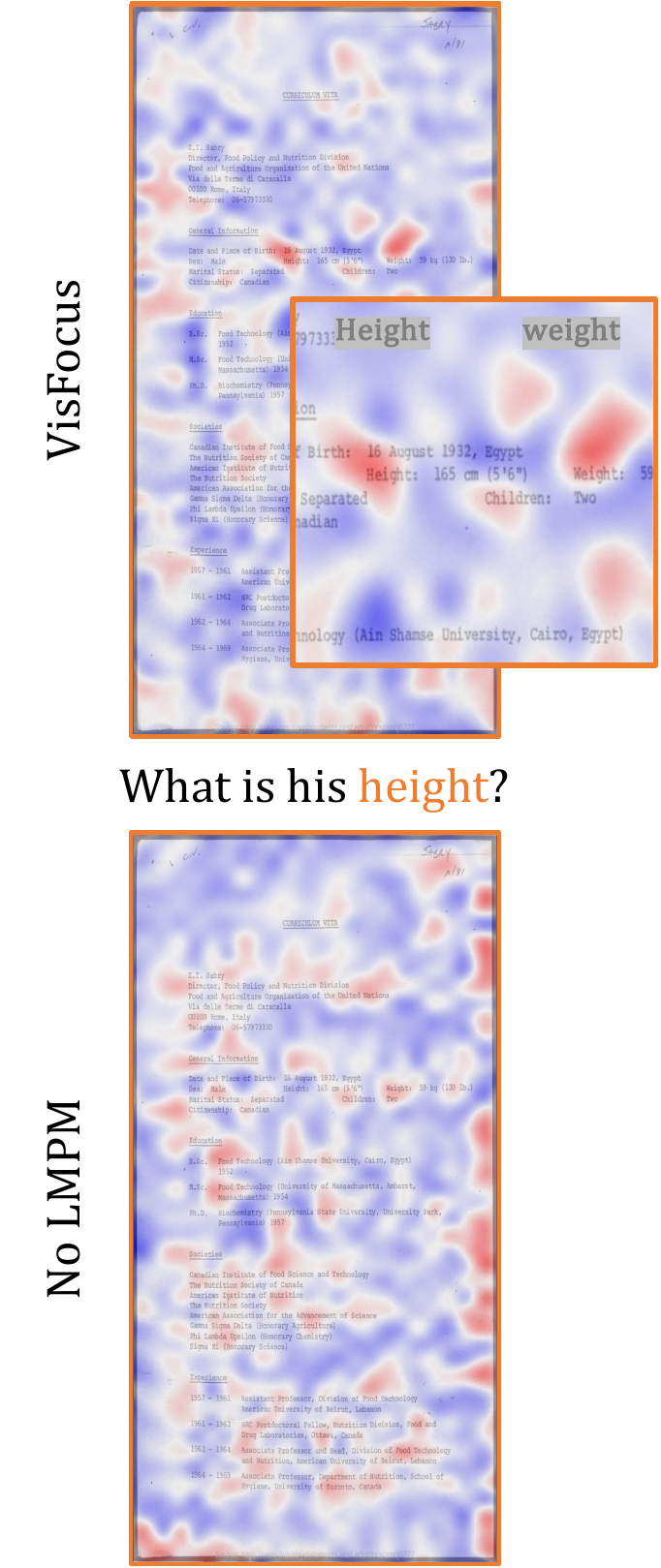}
        \caption{}
        \label{fig:supp_more_vis_3_5}
    \end{subfigure}
    \caption{\cref{fig:supp_more_vis_1} continued.}
    \label{fig:supp_more_vis_3}
\end{figure}

As discussed, our novel LMPM pre-training encourages the model to focus on relevant portions of the document, concerning the input query. This is particularly demonstrated by the learned attention maps within the last ViLMA layers (\cref{fig:supp_more_vis_1_1,fig:supp_more_vis_2,fig:supp_more_vis_3}), highlighting the correspondence between similar textual (query) and visual (context) tokens. The learned attention encompasses both  
 literal word-level alignment (`center' $\leftrightarrow$ `center' in \cref{fig:supp_more_vis_1_1}, `100' $\leftrightarrow$ `100' in \cref{fig:supp_more_vis_1_3}) and semantic relations (`birth' $\leftrightarrow$ `national' in \cref{fig:supp_more_vis_1_1}, `be' $\leftrightarrow$ `are' in \cref{fig:supp_more_vis_1_2}, `height' $ \leftrightarrow$ `weight' in \cref{fig:supp_more_vis_3_5}).

\section{OCR-Free vs. OCR-Based}
The OCR-free branch in document understanding aims to eliminate the need for external OCR systems, offering a more efficient standalone approach for processing document images. Consequently, OCR-free models' performance currently lags behind traditional OCR-based methods. This limitation stems from the absence of explicit textual information, which OCR-based models leverage as an additional input modality. \cref{table:ocr_based} compares the two branches. Despite the remaining performance gaps, OCR-based methods depend on external systems, which indirectly add model parameters and are pruned to error propagation and thus heavily rely on the quality of the OCR engines.

\renewcommand{\arraystretch}{1.2}
\setlength{\tabcolsep}{4pt}
\begin{table}
    \scriptsize    
    \begin{center}
        \caption{\textbf{Comparison with OCR-based methods on VQA benchmarks.} While the OCR-based approach still dominates in performance, the remaining gap with respect to OCR-free methods depends on the quality of external OCR engines which also implicitly add more parameters (\textit{P}*) and complexity to the system.}
        \vspace{-3mm}
        \label{table:ocr_based}
        \resizebox{\textwidth}{!}{
        \begin{tabular}{cl|c|cc}
            \toprule
            & \multirow{2}{*}{Method} & \multirow{2}{*}{\#params} & DocVQA & InfoVQA \\
            & & & \tiny{ANLS} & \tiny{ANLS} \\
            \hline & \\[-2ex]
            \multirow{5}{*}{\rotatebox{90}{OCR-based}}
             & LayoutLMV2-B\cite{huang2022layoutlmv3} & 200M + \textit{P}* & 78.1 & - \\
             & LayoutLMV2-L\cite{huang2022layoutlmv3} & 426M + \textit{P}* & 83.4 & - \\
             & LayoutLMV3\cite{huang2022layoutlmv3} & 794M + \textit{P}* & 83.4 & 45.1 \\
             & UDOP\cite{Tang2022UnifyingVTUDOP} & 794M + \textit{P}* & 84.7 & 47.4 \\
             & DocFormerV2-L\cite{appalaraju2023docformerv2AAAI} & 368M + \textit{P}* & \textbf{87.8} & \textbf{48.8} \\
            \hline & \\[-2ex]
            \multirow{5}{*}{\cellcolor{white}\rotatebox{90}{OCR-free}}
            & Dessurt\cite{davis2022dessurt} & 127M & 63.2 & - \\
            & Donut\cite{kim2021donut} & 176M & 67.5 & 11.6 \\
            & ScreenAI-B\cite{baechler2024screenai} & 670M & 50.7 & - \\
            & Pix2Struct-B\cite{lee2023pix2struct} & 282M & 72.1 & \textbf{38.2} \\
            & \cellcolor{gray!20}\AlgoNameNoSpace-B & \cellcolor{gray!20} 408M & \cellcolor{gray!20}\textbf{72.9}\gcol{\texttt{+}1.2} & \cellcolor{gray!20}31.9\gcol{\texttt{+}5.1} \\
            \bottomrule
            \end{tabular}
            }
        \end{center}
    \end{table}
  \vspace{-6mm}

\section{Qualitative Comparisons}
\begin{figure}[ht]
    \caption{\textbf{Qualitative comparison.} Further examples from the DocVQA validation set, demonstrating \AlgoNameNoSpace's ability to accurately answer questions on denser documents, compared to previous SoTA (Pix2Struct-B) and to our baseline}
    \centering
    \begin{subfigure}[t]{.75\textwidth}
        \centering
        \includegraphics[width=1\linewidth]{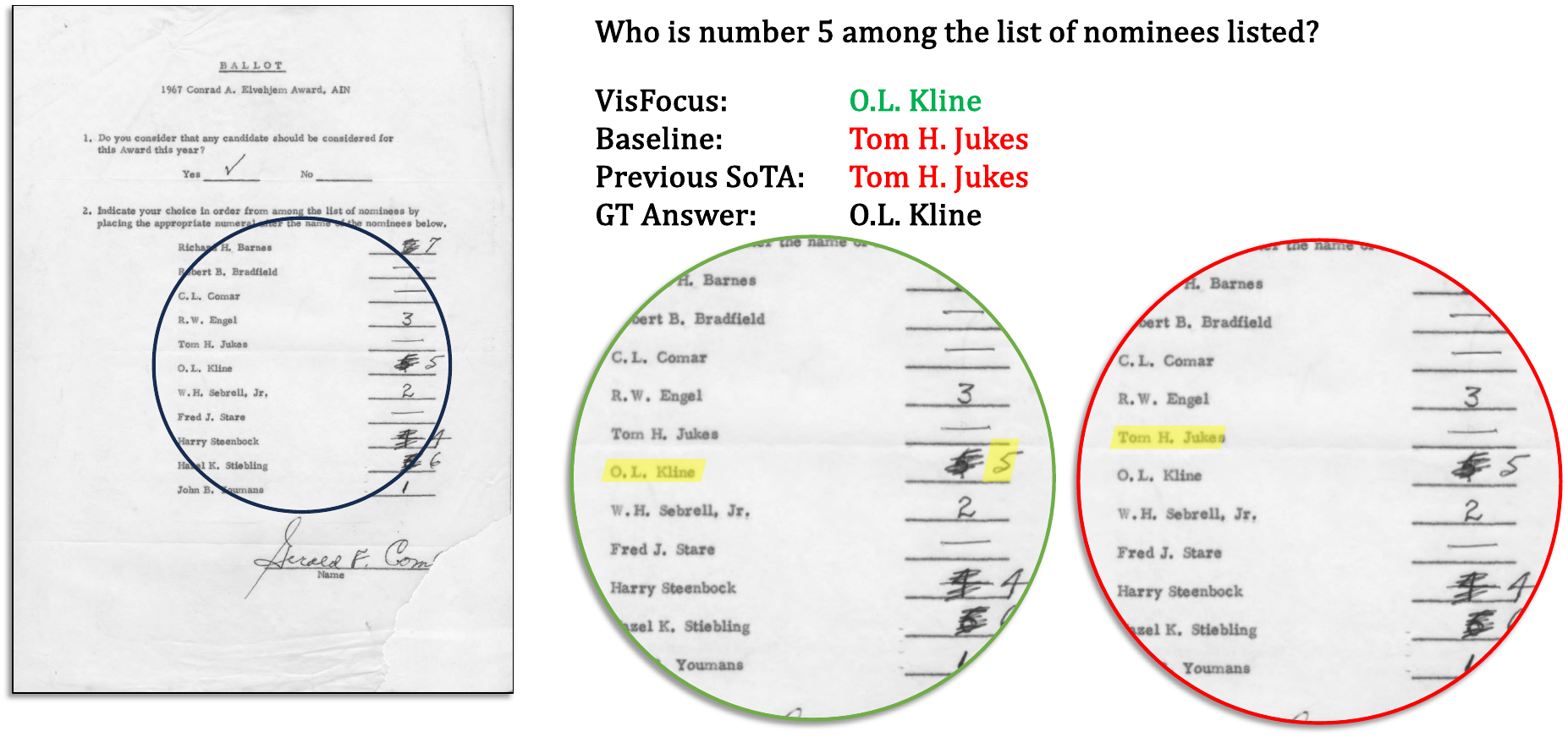}
    \end{subfigure}
    
    \vspace{1pt}
    \rule{\textwidth}{0.5pt}
    \vspace{1pt}

    \begin{subfigure}[t]{.75\textwidth}
        \centering
        \includegraphics[height=.5\linewidth]{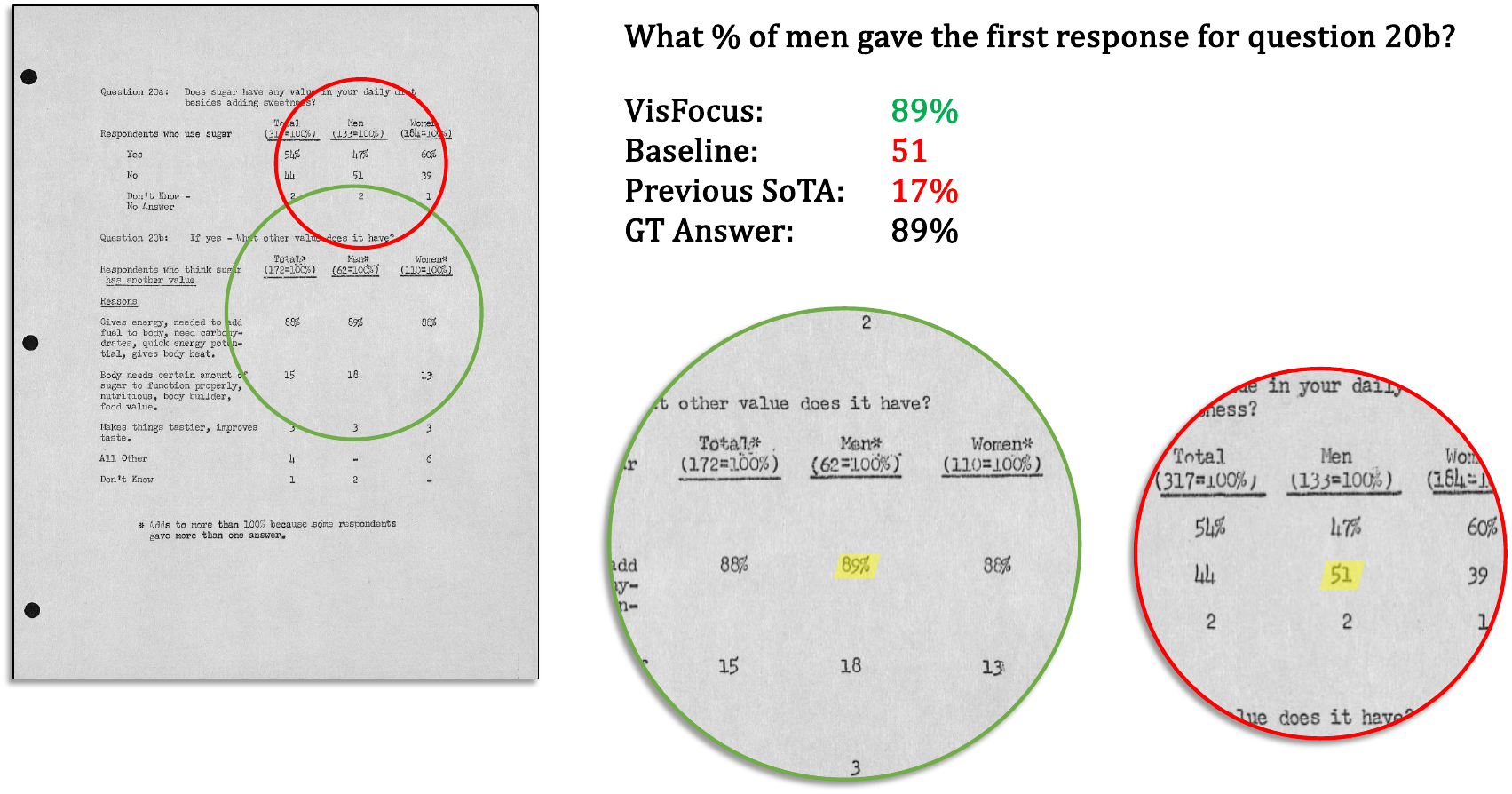}
    \end{subfigure}
  
    \vspace{1pt}
    \rule{\textwidth}{0.5pt}
    \vspace{1pt}

    \begin{subfigure}[t]{1\textwidth}
        \centering
        \includegraphics[height=.4\linewidth]{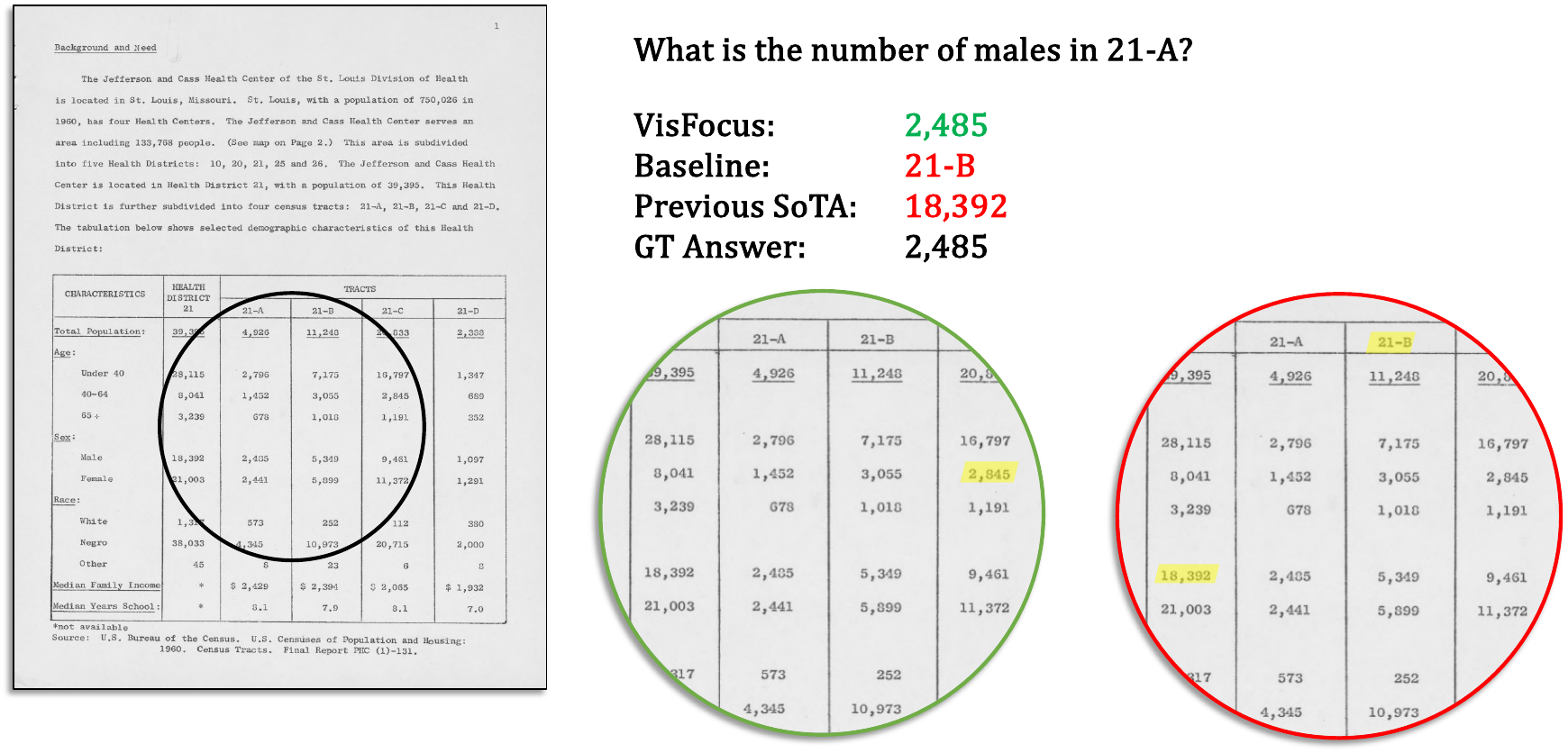}
    \end{subfigure}
    \label{fig:comp_1}
\end{figure}

\begin{figure}[ht]
  \centering
    \begin{subfigure}[t]{1\textwidth}
        \centering
        \includegraphics[width=1\linewidth]{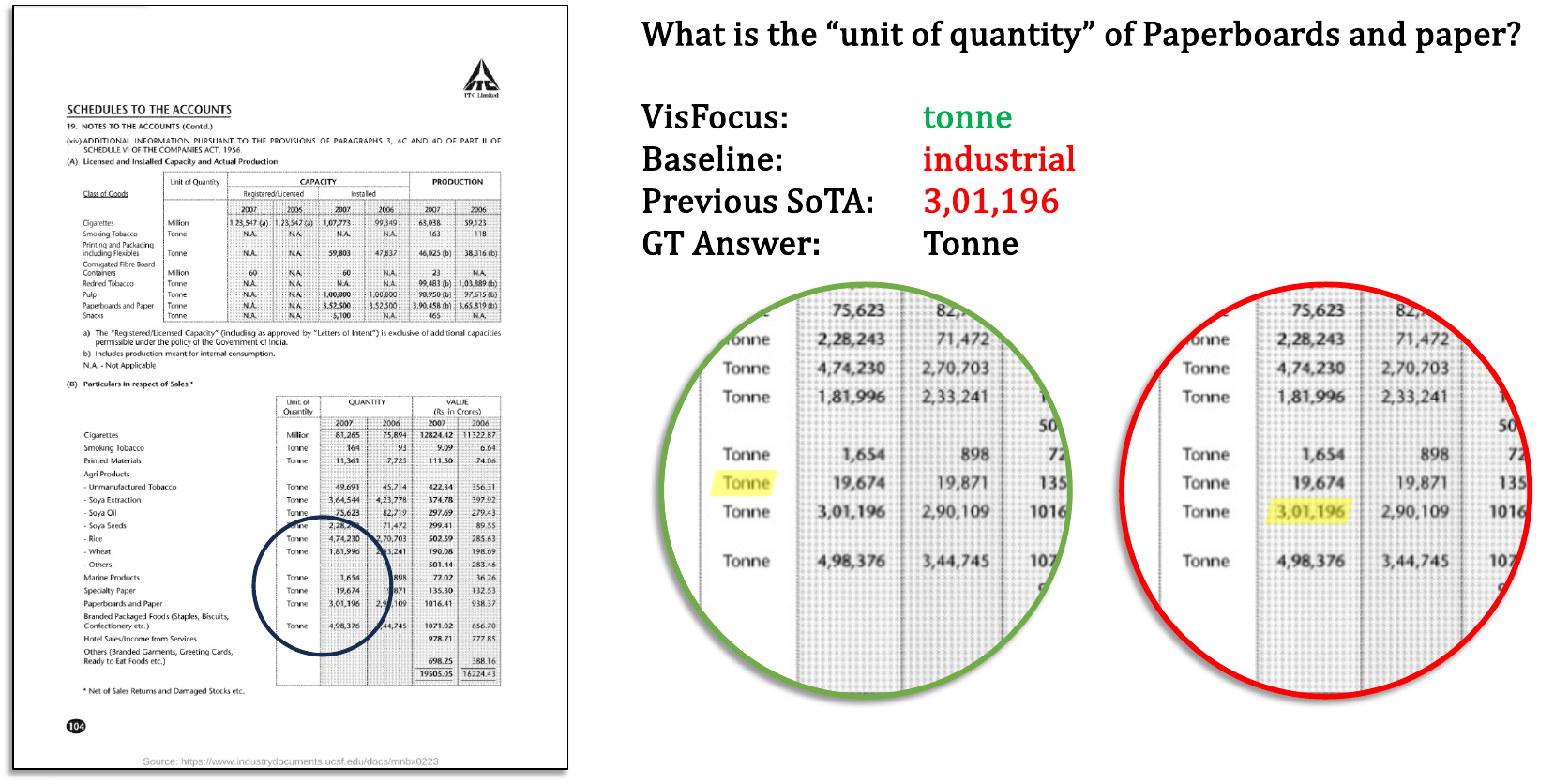}
    \end{subfigure}
    
    \vspace{1pt}
    \rule{\textwidth}{0.5pt}
    \vspace{1pt}

    \begin{subfigure}[t]{1\textwidth}
        \centering
        \includegraphics[width=1\linewidth]{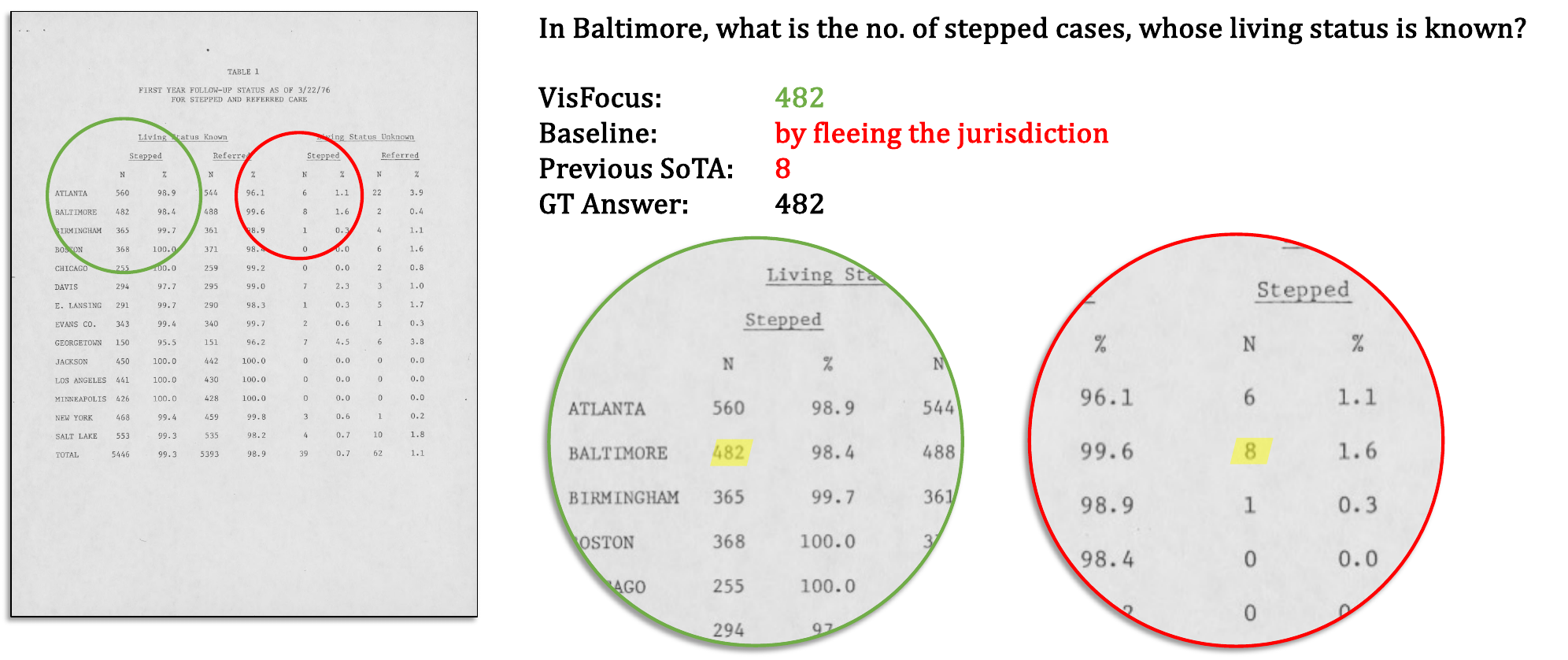}
    \end{subfigure}
    \caption{\cref{fig:comp_1} continued.}
    \label{fig:comp_2}
\end{figure}

\begin{figure}[ht]
    \centering
    \caption{\textbf{Qualitative comparison.} Further examples from ChartQA test set, demonstrating \AlgoNameNoSpace's ability to accurately answer questions on visual data such as charts and plots, compared to previous SoTA (Pix2Struct-B) and to our baseline. The last example shows a fail case for all of the compared methods.}
        \centering
        \includegraphics[height=.4\linewidth]{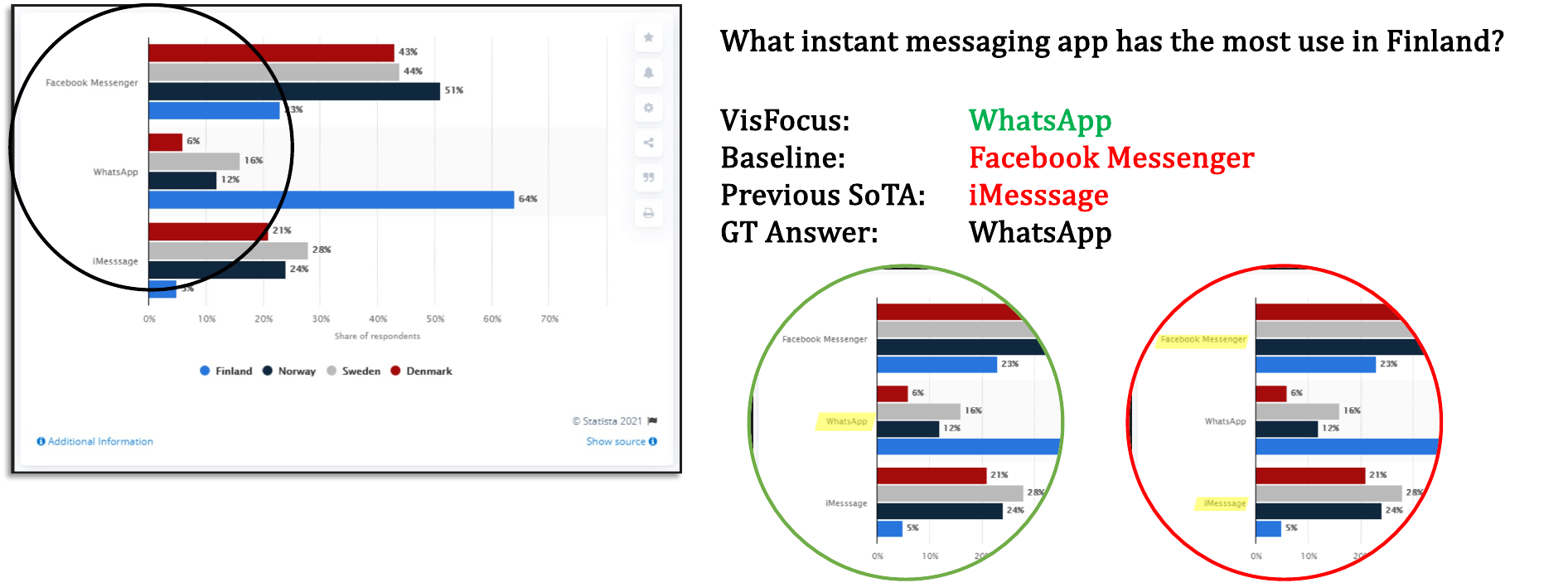}
    \vspace{1pt}
    \rule{\textwidth}{0.5pt}
    \label{fig:comp_3}
\end{figure}

\begin{figure}[H]
  \centering
    \begin{subfigure}[t]{1\textwidth}
        \centering
        \includegraphics[height=.35\linewidth]{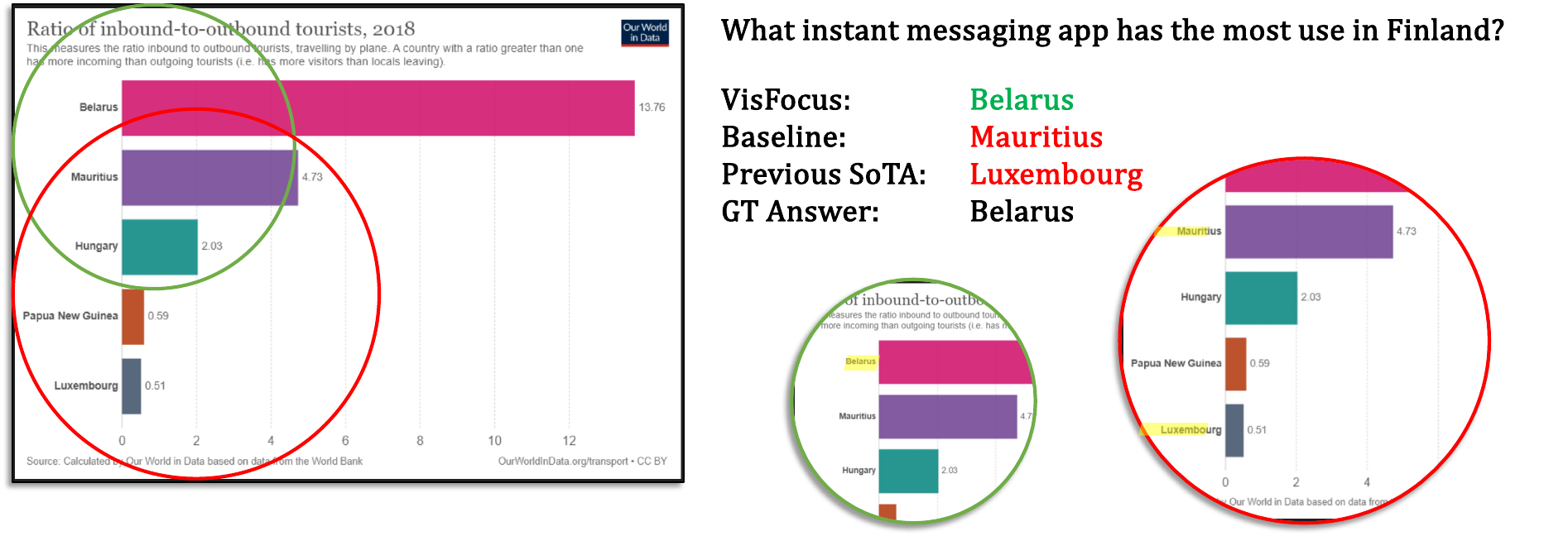}
    \end{subfigure}

    \vspace{1pt}
    \rule{\textwidth}{0.5pt}
    \vspace{1pt}

    \begin{subfigure}[t]{1\textwidth}
        \centering
        \includegraphics[height=.35\linewidth]{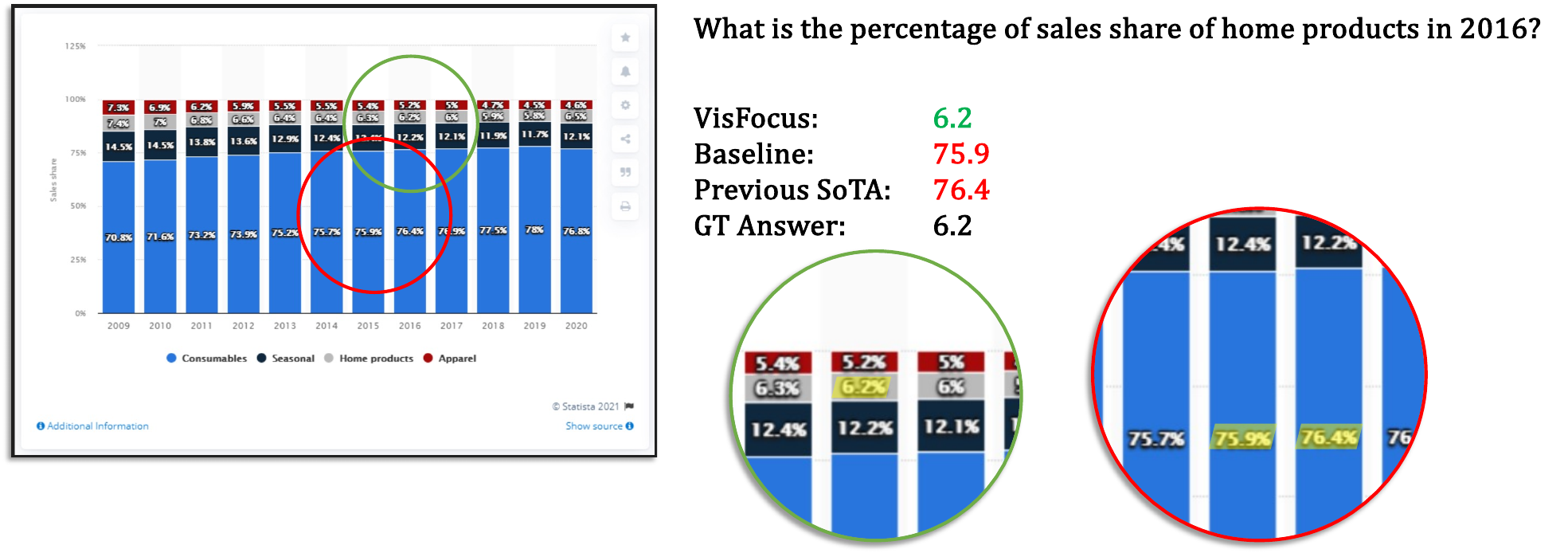}
    \end{subfigure}
    \caption{\cref{fig:comp_1} continued.}
    \label{fig:comp_4}
\end{figure}

\begin{figure}
    \centering
    \begin{subfigure}[t]{1\textwidth}
        \includegraphics[height=.5\linewidth]{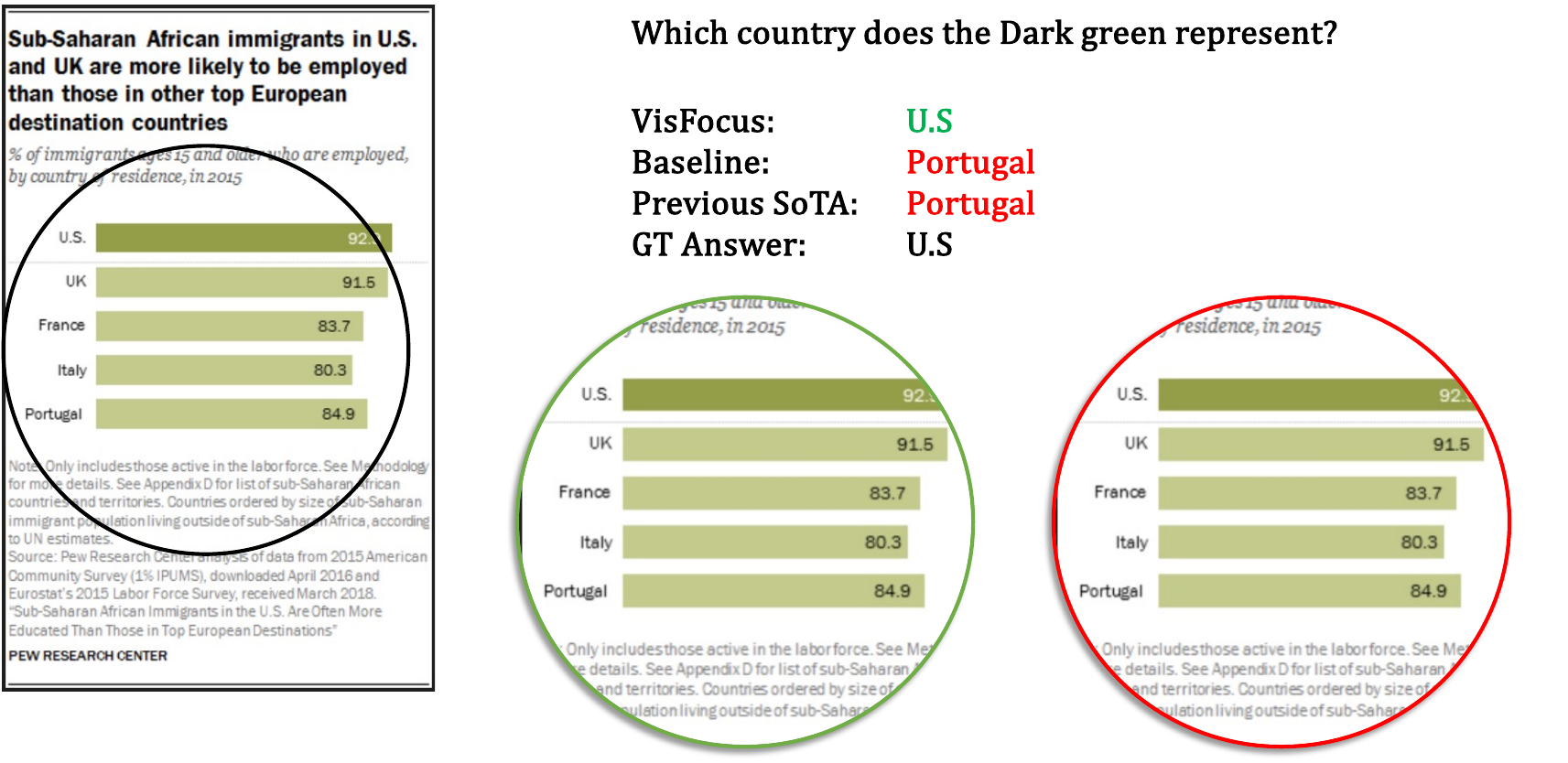}
    \end{subfigure}

    \vspace{1pt}
    \rule{\textwidth}{0.5pt}
    \vspace{1pt}

    \begin{subfigure}[t]{1\textwidth}
        \centering
        \includegraphics[height=.4\linewidth]{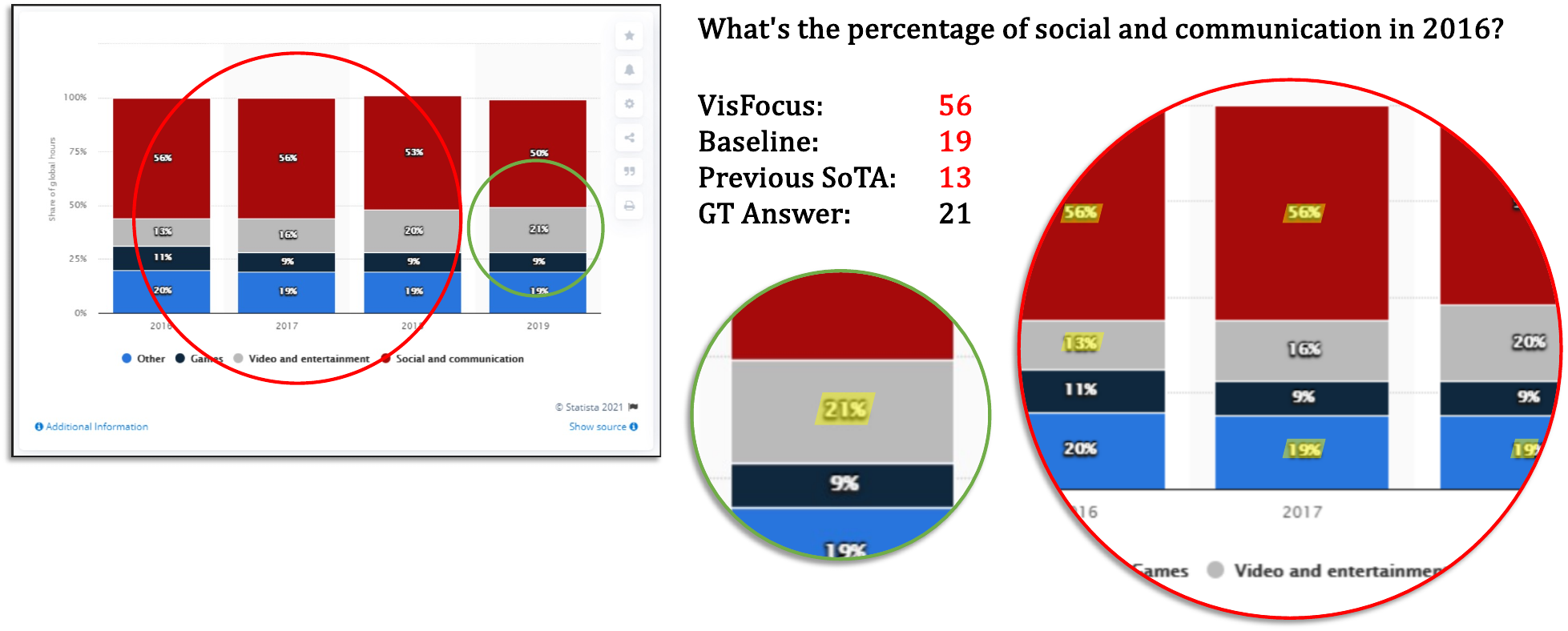}
    \end{subfigure}
    \caption{\cref{fig:comp_1} continued.}
    \label{fig:comp_5}
\end{figure}

\begin{figure}[ht]
    \centering
    \vspace{-5mm}
    \begin{subfigure}[ht]{.3\textheight}
        \centering
        \includegraphics[width=1\linewidth]{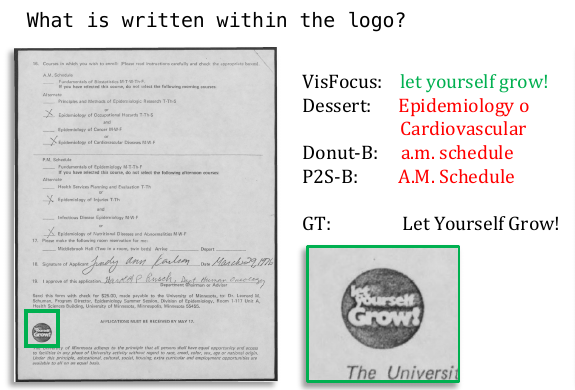}
        \caption{}
        \label{fig:pos_a}
    \end{subfigure}
    \vrule
    \begin{subfigure}[ht]{.3\textheight}
        \centering
        \includegraphics[width=1\linewidth]{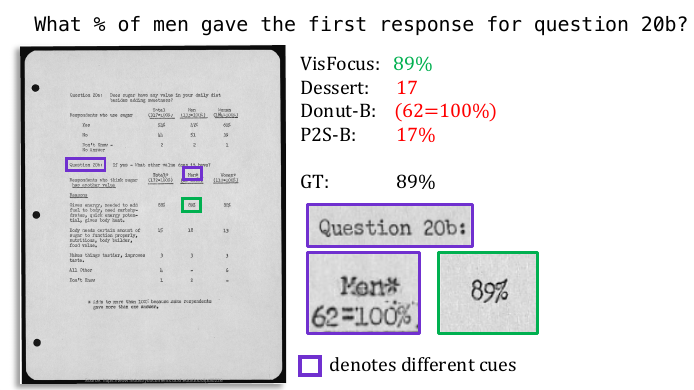}
        \caption{}
        \label{fig:pos_b}
    \end{subfigure}
    
    \vspace{1pt}
    \rule{\textwidth}{0.5pt}
    \vspace{1pt}
    
    \begin{subfigure}[ht]{.3\textheight}
        \centering
        \includegraphics[width=1\linewidth]{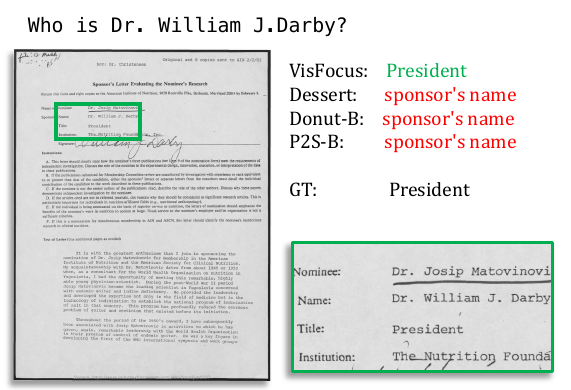}
        \caption{}
        \label{fig:pos_c}
    \end{subfigure}
    \vrule
    \begin{subfigure}[ht]{.3\textheight}
        \centering
        \includegraphics[width=1\linewidth]{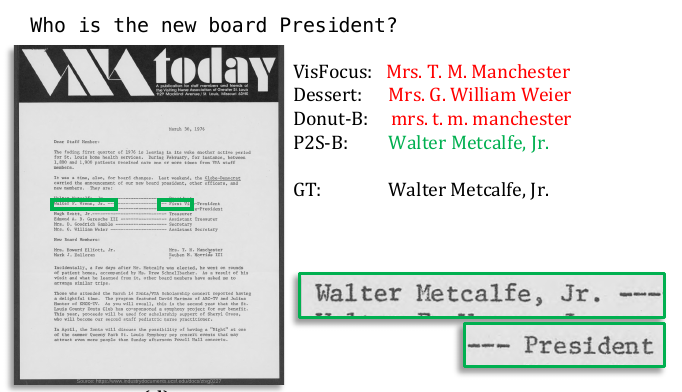}
        \caption{}
        \label{fig:pos_d}
    \end{subfigure}
    \caption{\textbf{Qualitative comparison.} Further success cases of \AlgoNameNoSpace, compared to the failures of other OCR-free methods: Dessurt, Donut and Pix2Struct-B.}
    \label{fig:pos}
\end{figure}

\begin{figure}[ht]
    \centering
    \vspace{-5mm}
    \begin{subfigure}[ht]{.2\textheight}
        \centering
        \includegraphics[width=1\linewidth]{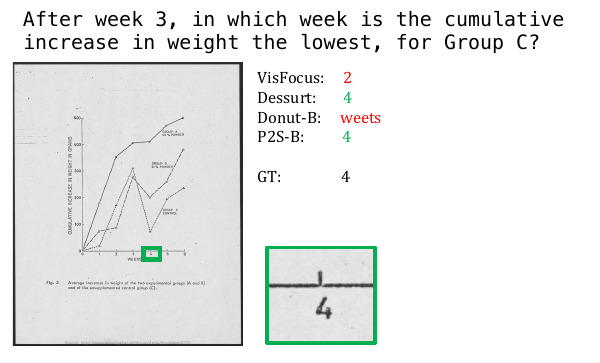}
        \caption{}
        \label{fig:neg_a}
    \end{subfigure}
    \vrule
    \begin{subfigure}[ht]{.2\textheight}
        \centering
        \includegraphics[width=1\linewidth]{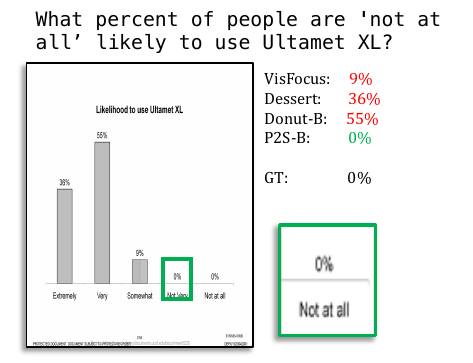}
        \caption{}
        \label{fig:neg_b}
    \end{subfigure}
        \vrule
    \begin{subfigure}[ht]{.2\textheight}
        \centering
        \includegraphics[width=1\linewidth]{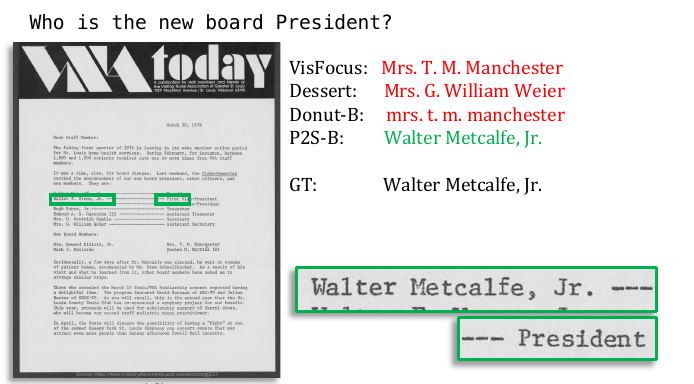}
        \caption{}
        \label{fig:neg_c}
    \end{subfigure}
        \caption{\textbf{Qualitative comparison.} Failure examples of cases where \AlgoName fails and other OCR-free methods: Dessurt, Donut and  Pix2Struct-B succeed better.}
    \label{fig:neg}
\end{figure}

\cref{fig:comp_1,fig:comp_2,fig:comp_3,fig:comp_4,fig:comp_5} provide additional examples where \AlgoNameNoSpace-B excels in comparison to our baseline and Pix2Strcut-B and \cref{fig:pos} extends the comparison to other OCR-free methods: Dessurt, Donut and Pix2Strcut-B. It can be seen that all but \AlgoNameNoSpace often predict wrong answers, extracted from somewhere in the document. This implies on the lack of focusing, as discussed in \ref{apdx: attention maps}, which leads to extraction of unrelated information and in turn to wrong predictions. \cref{fig:neg} shows fail cases of \AlgoNameNoSpace compared to other OCR-free methods.

\section{\AlgoName on Zero-shot Key-Value Extraction}
\AlgoName is originally designed for prompt-related document VQA tasks, but can be adapted to demonstrate its versatility on other document understanding tasks. One such task is key-value extraction, which can be reformulated as a prompt-related task. This reformulation allows leveraging \AlgoNameNoSpace's capabilities beyond its original design scope.

\begin{wrapfigure}{r}{0.4\textwidth}
\vspace{-3mm}

\renewcommand{\arraystretch}{1.2}
\setlength{\tabcolsep}{8pt}
    \begin{center}
    \vspace{-3em}
        \captionof{table}{\textbf{Comparison of zero-shot Relaxed KV Extraction task on FUNSD dataset.} We report ANLS on the test set, applying the reformulated KV task.}
        \label{table:zero-shot}
        \vspace{.75em}
        \begin{tabular}{l|c}
            \toprule
            Method & ANLS \\
            \hline  \\[-2ex]
            Donut  & 58.9 \\
            VisFocus-S  & \textbf{60.2} \gcol{\texttt{+}1.3} \\
            \midrule
            Pix2Struct-B & 62.7 \\
            VisFocus-B  & \textbf{63.4} \gcol{\texttt{+}0.7} \\
            \bottomrule
        \end{tabular}
    \end{center}

\end{wrapfigure}
\begin{figure}[h]
    \centering
    \includegraphics[width=.9\linewidth]{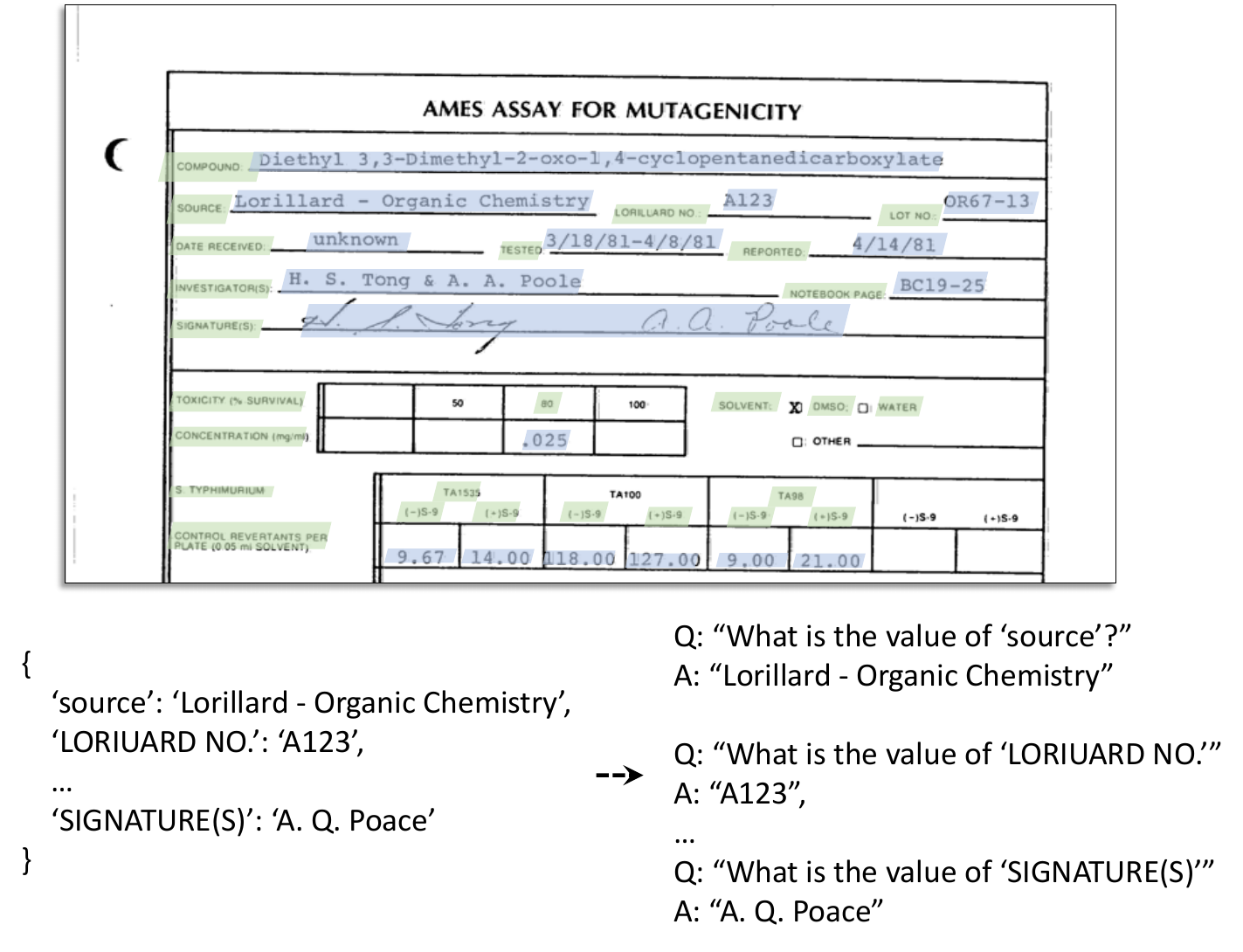}
    \caption{\textbf{Visualization of the Relaxed KV Extraction.} We re-define the key-value extraction as a prompt-based task to apply zero-shot on VQA fine-tuned models. \fcolorbox{lightGreen}{lightGreen}{\textcolor{lightGreen}{-}} and \fcolorbox{lightBlue}{lightBlue}{\textcolor{lightBlue}{-}} denote keys and values respectively.}
    \label{fig:rkv_vis}
\end{figure}

To accomplish this, the key-value extraction task is reframed using a prompt template:
``\textit{What is the value of <key>?}''. where \textit{<key>} is some key in the form/reciept (\cref{fig:rkv_vis}). We refer to this task as \textit{relaxed KV extraction}, since one should know a key in the document, and prompt it.
We evaluate our proposed task on the FUNSD dataset \cite{jaume2019funsd}, using DocVQA-finetuned checkpoints of \AlgoName and previous works, and report superior performance in the zero-shot setting (\cref{table:zero-shot}). This evaluation strategy demonstrates the flexibility of prompt-based models like \AlgoName and explores their potential for tackling diverse document understanding tasks through clever task reformulation.

\section{Datasets and Hyperparameters}
In this section we present in detail every benchmark and dataset used in our work. For more pre-training and fine-tuning details see \cref{table:pt_hp,table:ft_hp}.

\setlength{\tabcolsep}{4pt}
\begin{table}[h]
    \scriptsize
    \begin{center}
        \caption{\textbf{Model hyper-parameters for pre-training.} We use AdamW\cite{DBLP:journals/corr/abs-1711-05101} optimizer and Cosine Annealing\cite{DBLP:journals/corr/LoshchilovH16a} scheduler. We train on 8 A100 GPUs. `$*$' denotes early stopping. All reported numbers apply for all our model variants.}
        \label{table:pt_hp}
        \resizebox{\textwidth}{!}{
        \begin{tabular}{lcccccc}
            \toprule
            PT Stage & \#steps & Batch Size & Base LR & Image Resolution \\
            \midrule
            LtR  & $200K^*$ & $32$ & \multirow{2}{*}{$1e-4$} & \multirow{2}{*}{$1536\times 768$} \\
            LMPM & $400K$ & $48$ &  \\
            \bottomrule
        \end{tabular}
        }
    \end{center}
    \end{table}
\

\setlength{\tabcolsep}{4pt}
\begin{table}[h]
    \scriptsize    
    \begin{center}
        \caption{\textbf{Model hyper-parameters for fine-tuning.} Same as in pre-training, in all our experiments we adopt AdamW\cite{DBLP:journals/corr/abs-1711-05101} optimizer, Cosine Annealing\cite{DBLP:journals/corr/LoshchilovH16a} scheduler, early-stopping and train on 8 A100 GPUs.}
        \label{table:ft_hp}
        \resizebox{\textwidth}{!}{
        \begin{tabular}{clcccccc}
            \toprule
            & Dataset & \#Steps & Batch Size & Base LR &  Image Resolution \\
            \midrule
            \multirow{5}{*}{\rotatebox{90}{\AlgoNameNoSpace-S}}
            & DocVQA  & $15K$ & $72$ & $1e-4$ &   \multirow{5}{*}{$1536\times 768$}\\
            & InfoVQA & $15K$ & $32$ & $5e-5$  \\
            & ChartQA & $15K$ & $72$ & $2e-4$  \\
            & OCR-VQA & $50K$ & $64$ & $5e-5$  \\
            & AI2D    & $30K$ & $512$ & $1e-4$ \\
            \midrule
            \multirow{5}{*}{\rotatebox{90}{\AlgoNameNoSpace-B}}
            & DocVQA  & $15K$ & $72$ & $1e-4$ &  \multirow{5}{*}{$1536\times 768$}\\
            & InfoVQA & $15K$ & $144$ & $1e-4$ \\
            & ChartQA & $15K$ & $72$ & $2e-4$ \\
            & OCR-VQA & $50K$ & $144$ & $1e-4$ \\
            & AI2D & $30K$ & $32$ & $5e-5$ \\
            \bottomrule
        \end{tabular}
        }
    \end{center}
    \end{table}
\setlength{\tabcolsep}{1.4pt}

\subsection{Pre-training Data}
For pretraining data, we utilize the IDL-OCR dataset\cite{biten2022ocr}, comprising 26M document pages accompanied by corresponding raster-scan OCR outputs. In \cref{table:pretraining} we compare our pre-training data with previous methods. \cite{davis2022dessurt,lee2023pix2struct} create different labels for documents, whereas we employ only the OCR text, similar to \cite{kim2021donut}. Dessurt collects textual data to create synthetic documents using open-sourced fonts. It also re-renders IIT-CDIP\cite{harley2015icdar,10.1145/1148170.1148307} and FUNSD\cite{jaume2019funsd} with different fonts and layouts, while Pix2Struct scrape the web to generate structured representations of documents (HTML DOMs). Pre-training our model with text-oriented approaches, further provides an advantage for our method when dealing with dense documents of many words.%

\setlength{\tabcolsep}{4pt}
\begin{table}[h]
    \begin{center}
        \caption{Pre-traing Data. Comparison with previous OCR-Free methods. ``I'' is denoted as the IIT-CDIP dataset~\cite{harley2015icdar}, `Form',`Handwriting', and `Wiki' are synthetic datasets presented in \cite{davis2022dessurt}. OCR refers to raster-scan order.}
        \label{table:pretraining}
        \begin{tabular}{lccc}
            \toprule
            & Pre-training Datasets & Annotations & \#Samples\\
            \midrule
            Dessurt \cite{davis2022dessurt} & I+Form+Handwriting+Wiki & OCR & not reported \\
            Donut \cite{kim2021donut} & I+SynthDog\cite{kim2021donut} & OCR & 13.5M \\
            Pix2Struct \cite{lee2023pix2struct} & C4 \cite{2019t5} &  HTML DOMs + OCR & 80M \\
            \midrule
            \AlgoNameNoSpace & IDL-OCR\cite{biten2022ocr} & OCR & 25.6M \\
            \bottomrule
        \end{tabular}
    \end{center}
    \end{table}
\setlength{\tabcolsep}{1.4pt}

\subsection{Downstream Tasks}
Here we provide technical details about the downstream datasets we experimented with and some bottom line results.

\subsubsection{DocVQA\cite{mathew2021docvqa}}
is an open-ended VQA dataset consists of various types of scanned documents. It is a subset of IDL corpus, consists of \(\sim\)$14k$ document images and \(\sim\)$40k$ questions. We use the ANLS metric and report a boost of $\mathbf{+1.2}$ on the test split over the baseline, and $\mathbf{+0.8}$ over Pix2Struct-B, which is the current state-of-the-art on small OCR-free models.

\subsubsection{InfographicsVQA (InfoVQA)\cite{mathew2022infographicvqa}} contains various infographics with annotations for questions that demand reasoning across text, layout, graphics, and data visualizations. It consists of \(\sim\)$5k$ images and \(\sim\)$30k$ questions. Since \AlgoName was trained to encode the question with respect the question, and given that InfoVQA has more visual than textual content, along with complex numerical reasoning, its performance drops and becomes less competitive. However, our approach still beats the baseline by $\mathbf{+5.1}$ points.

\vspace{-3mm}
\subsubsection{ChartQA\cite{masry-etal-2022-chartqa}}
is a large-scale benchmark dataset designed to evaluate models' ability to answer complex questions about charts, requiring both visual and logical reasoning. It consists of $9.6K$ human-written questions and $23.1K$ generated questions based on summaries. We follow previous works and report average Relaxed Accuracy (RA) of each split. Even though \AlgoName is not trained on structure-related tasks, it achieves an improvement of $\mathbf{+4.6}$ over the baseline, and exceeding previous works.

\subsubsection{OCR-VQA\cite{mishraICDAR19}} is a large-scale dataset of \(\sim\)$200k$ book cover images and $1M$ questions. The task requires high skills of reading text. We report Exact Match (EM) on the test set and outperform our baseline by $\mathbf{+3.1}$ and $\mathbf{+0.6}$ points over the baseline and Pix2Struct-B, respectively.

\subsubsection{AI2 Diagrams (AI2D)\cite{Kembhavi2016ADI}} consists of \(\sim\)$5K$ grade-school science diagrams, corresponding multiple-choice questions testing comprehension and reasoning about the diagrams. \AlgoNameNoSpace-B achieves a $\mathbf{+2.2}$ boost over the baseline and $\mathbf{+6.9}$ compared to Pix2Struct-B, the prior state-of-the-art model in our setting.

\subsection{Prompt Encoding}
To quantify the impact of the prompt encoding, we conduct an ablation study on several different text encoding techniques, involving both context-aware encoding (T5 encoder\cite{2019t5}, TinyBERT\cite{jiao2019tinybert}) and independent learned token embeddings (\cref{table:abl_text_encoder}). It can be seen that designated T5 based encoders perform best, possibly due to the alignment with the T5 language model cascaded to the vision encoder. Reducing its size by about 65\% decreases the DocVQA ANLS by merely $0.2$. This motivates further research of smaller variants for T5 to be utilized in our framework.
\setlength{\tabcolsep}{4pt}
\begin{table}[h]
    \begin{center}
        \caption{\textbf{Prompt Encoding Ablation Study.} `Embedding' denotes independent token embedding and `shared' as the \AlgoNameNoSpace's LM encoder. `shared' uses the T5 encoder for encoding the textual prompt in addition to the visual features. `Cross-modal grad.' computes gradients from both paths.}
        \label{table:abl_text_encoder}
        \begin{tabular}{lccc}
            \toprule
            \multirow{2}{*}{Prompt Encoder} & $\Delta$ \#Params & DocVQA       & ChartQA   \\
                                            & (frozen)          & \tiny{ANLS}  & \tiny{RA}        \\
            \midrule
            Embedding                                   & - & 71.4 & 55.4 \\
            TinyBERT\cite{jiao2019tinybert}             & 8M & 71.3    & 56.2    \\
            T5-Base Enc. (Copy of the learnt LM)    & - & 71.7 & 56.0 \\
            T5-Small Enc.                               & 39M & 72.0 & 56.3 \\
            \textbf{T5-Base Enc.}                & \textbf{113M} & \textbf{72.2} & \textbf{57.1} \\
            \bottomrule
        \end{tabular}
    \end{center}
    \vspace{-5mm}
    \end{table}
\setlength{\tabcolsep}{1.4pt}

\clearpage

\clearpage

\clearpage

\end{document}